\documentclass{vgtc}              

\onlineid{1436}
\onlineid{1015}

\vgtccategory{Research}

\vgtcpapertype{Algorithm/Technique}

\title{Topology Aware Neural Interpolation of Scalar Fields}

\author{%
  \topoinvis{Mohamed Kissi\thanks{\topoinvis{CNRS, Sorbonne Université -
\href{mailto:mohamed.kissi@sorbonne-universite.fr}{mohamed.kissi@sorbonne-universite.fr}}},
  Keanu Sisouk\thanks{\topoinvis{CNRS, Sorbonne Université - \href{mailto:keanu.sisouk@sorbonne-universite.fr}{keanu.sisouk@sorbonne-universite.fr}}},
  Joshua A. Levine\thanks{\topoinvis{University of Arizona - \href{mailto:josh@cs.arizona.edu}{josh@cs.arizona.edu}}}, and
  Julien Tierny\thanks{\topoinvis{CNRS, Sorbonne Université - \href{mailto:julien.tierny@sorbonne-universite.fr}{julien.tierny@sorbonne-universite.fr}}}
  }
}

\authorfooter{
  \item
    Mohamed Kissi, Keanu Sisouk, and Julien Tierny are with the CNRS and
Sorbonne University.
  	E-mail: \{firstname.lastname\}@sorbonne-universite.fr
  \item
    Joshua A. Levine is with the University of Arizona.
    E-mail: josh@cs.arizona.edu
}

\abstract{This is an abstract, line1.\\
line 2.\\
line 3.\\
line 4.\\
line 5.\\
line 6.\\
line 7.\\
line 8.\\
line 9.\\
line 10.\\
line 11.\\
line 12.}

\abstract{%
This paper presents a neural scheme
for the topology-aware
interpolation of time-varying scalar fields.
Given a time-varying sequence of
persistence diagrams, along with a sparse temporal sampling of the corresponding
scalar fields, denoted as \emph{keyframes},
our
\revision{interpolation approach aims at \emph{``inverting''}}
the \emph{non-keyframe} diagrams to
produce plausible estimations of the corresponding, missing
data.
\revision{For this, we rely on a neural architecture which}
learns the relation from a time
value to the corresponding scalar field, based on the keyframe
examples, and reliably extends this relation to the non-keyframe time steps.
We show how augmenting this architecture with specific
\revision{topological}
losses
\revision{exploiting the input diagrams}
both improves the geometrical and topological reconstruction of the
non-keyframe time steps.
At query time, given an input time
value for which an interpolation
is desired, our approach instantaneously produces an output, via  a
single propagation of the time input through the network.
Experiments
interpolating 2D and 3D time-varying
datasets
\revision{show our approach superiority,}
both in terms of data and topological fitting, with
regard to
reference
interpolation schemes. Our implementation is available at \href{https://github.com/MohamedKISSI/Topology-Aware-Neural-Interpolation-of-Scalar-Fields.git}{this GitHub link}.}


\teaser{
  \centering
  \includegraphics[width=\linewidth]{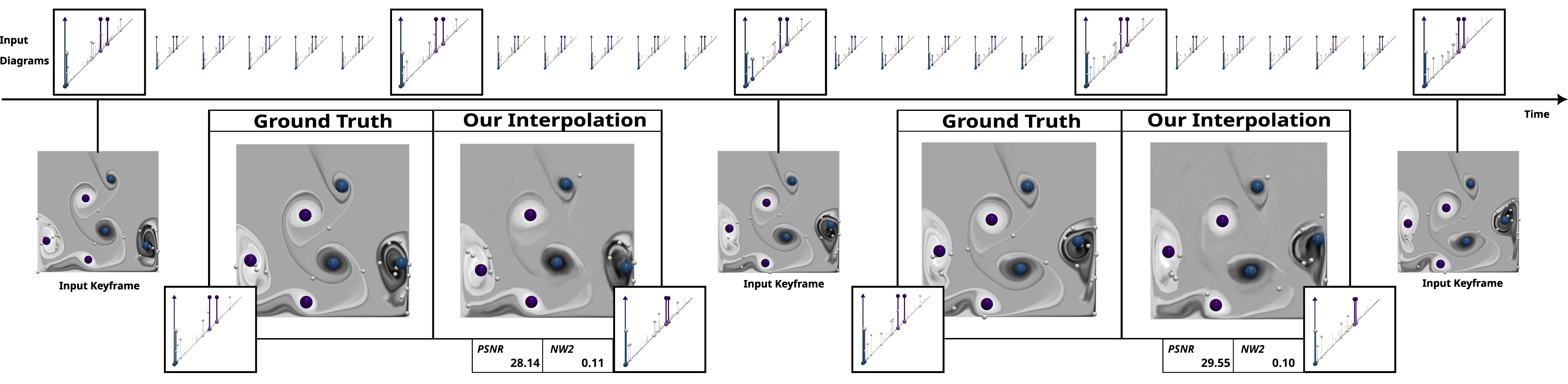}
  \caption{Given a time-varying sequence of
  persistence diagrams (top),
along
with a sparse set of corresponding \emph{keyframe} scalar fields (``Input
Keyframe'' labels),
our
\revision{interpolation approach aims at \emph{``inverting''}}
the
\emph{non-keyframe} diagrams and
\revision{generating}
plausible
estimations of the missing, non-keyframe scalar fields (``Our Interpolation'' 
labels)
enabling their
visualization and analysis.}
  \label{fig:teaser}
}

\graphicspath{{figs/}{figures/}{pictures/}{images/}{./}} 

\usepackage{times}                     

\usepackage{tabu}                      
\usepackage{booktabs}                  
\usepackage{lipsum}                    
\usepackage{mwe}                       

\usepackage{mathptmx}                  
\usepackage{amsmath}
\usepackage{amssymb}
\usepackage{calrsfs}
\usepackage{eucal}
\usepackage{enumitem}
\usepackage{mathtools}
\usepackage{algorithm}
\usepackage{algorithmicx}
\usepackage{algpseudocode}

\usepackage{multirow}
\usepackage{hhline}
\usepackage{colortbl}
\usepackage{adjustbox}
\usepackage{hyperref}
\usepackage{xcolor}

\DeclareMathAlphabet{\pazocal}{OMS}{zplm}{m}{n}
\SetMathAlphabet\pazocal{bold}{OMS}{zplm}{bx}{n}

\begin{document}



\renewcommand{\mathcal}[1]{\pazocal{#1}}

\newcommand{\surface}{S}

\newcommand{\keyframes}{\mathcal{F}}
\newcommand{\diagramSequence}{\mathcal{P}}

\newcommand{\cubicalComplex}{\mathcal{C}}

\newcommand{\signal}{\mathcal{S}}
\newcommand{\noise}{\mathcal{N}}
\newcommand{\simplex}{\sigma}
\newcommand{\domain}{\mathcal{K}}
\newcommand{\numberOfVertices}{n_v}
\newcommand{\numberOfSimplices}{n_\simplex}
\newcommand{\dataVector}{v}
\newcommand{\dataVectorSpace}{\mathcal{V}}
\newcommand{\filtration}{\mathcal{F}}
\newcommand{\persistenceMap}{\mathcal{P}}
\newcommand{\energy}{\mathcal{E}}
\newcommand{\loss}{\mathcal{L}}
\newcommand{\range}{\mathbb{R}}
\newcommand{\sublevelset}[1]{#1^{-1}_{-\infty}}
\newcommand{\superlevelset}[1]{#1^{-1}_{+\infty}}
\newcommand{\Star}{St}
\newcommand{\Link}{Lk}
\newcommand{\diagram}{\mathcal{D}}
\newcommand{\target}{\diagram_T}
\newcommand{\complexity}{\mathcal{O}}

\newcommand{\face}{\tau}
\newcommand{\lowerlink}{\Link^{-}}
\newcommand{\upperlink}{\Link^{+}}
\newcommand{\Index}{\mathcal{I}}
\newcommand{\offset}{o}
\newcommand{\Natural}{\mathbb{N}}
\newcommand{\criticalSet}{\mathcal{C}}

\newcommand{\wasserstein}[1]{\mathcal{W}_{#1}}
\newcommand{\normalizedWasserstein}[1]{\mathcal{NW}_{#1}}
\newcommand{\projection}{\Delta}
\newcommand{\hierarchy}{\mathcal{H}}
\newcommand{\decimation}{D}
\newcommand{\xDimD}{L_x^\decimation}
\newcommand{\yDimD}{L_y^\decimation}
\newcommand{\zDimD}{L_z^\decimation}
\newcommand{\xDim}{L_x}
\newcommand{\yDim}{L_y}
\newcommand{\zDim}{L_z}
\newcommand{\Grid}{\mathcal{G}}
\newcommand{\GridD}{\mathcal{G}^\decimation}
\newcommand{\x}{\phantom{x}}
\newcommand{\Mod}{\;\mathrm{mod}\;}
\newcommand{\NN}{\mathbb{N}}
\newcommand{\forwardIntegralLine}{\mathcal{L}^+}
\newcommand{\backwardIntegralLine}{\mathcal{L}^-}
\newcommand{\triangulationOp}{\phi}
\newcommand{\decimationOp}{\Pi}
\newcommand{\isovalue}{w}
\newcommand{\persistence}{p}
\newcommand{\pointMetric}{d}
\newcommand{\diagramSet}{\mathcal{S}_\mathcal{D}}
\newcommand{\diagramSpace}{\mathbb{D}}
\newcommand{\jointree}{\mathcal{T}^-}
\newcommand{\splittree}{\mathcal{T}^+}
\newcommand{\mergetree}{\mathcal{T}}
\newcommand{\tree}{\mergetree}
\newcommand{\depth}{d}
\newcommand{\mergetreeSet}{\mathcal{S}_\mathcal{T}}
\newcommand{\branchset}{\mathcal{S}_\mathcal{B}}
\newcommand{\branchspace}{\mathbb{B}}
\newcommand{\mergetreeSpace}{\mathbb{T}}
\newcommand{\editdistance}{D_E}
\newcommand{\wassersteinTree}{W^{\mergetree}_2}
\newcommand{\distanceSequence}{d_S}
\newcommand{\branchtree}{\mathcal{B}}
\newcommand{\branchtreeSet}{\mathcal{S}_\mathcal{B}}
\newcommand{\branchtreeSpace}{\mathbb{B}}
\newcommand{\forest}{\mathcal{F}}
\newcommand{\sequenceSpace}{\mathbb{S}}
\newcommand{\forestMatrix}{\mathbb{F}}
\newcommand{\treeMatrix}{\mathbb{T}}
\newcommand{\normalizedLocation}{\mathcal{N}}
\newcommand{\geodesictree}{\mathcal{G}}
\newcommand{\dummyVector}{\mathcal{V}}
\newcommand{\geodesictreeVec}{g}
\newcommand{\geodesicAxis}{\mathcal{A}}
\newcommand{\directionVector}{\mathcal{V}}
\newcommand{\geodesicdiagram}{\mathcal{G}^{\diagram}}
\newcommand{\reconstructionError}{E_{L_2}}
\newcommand{\pcaBasis}{B_{\mathbb{R}^d}}
\renewcommand{\pcaBasis}{B}
\newcommand{\origin}{o_b}
\newcommand{\sizeEncoding}{n_e}
\newcommand{\sizeDecoding}{n_d}
\newcommand{\linearTransformation}{\psi}
\newcommand{\unitTransformation}{\Psi}
\renewcommand{\origin}{o}
\newcommand{\bdtOrigin}{\mathcal{O}}
\newcommand{\activation}{\sigma}
\newcommand{\validBDT}{\gamma}
\newcommand{\mtPgaBasis}{B_{\branchtreeSpace}}
\newcommand{\mtPgaError}{E_{\wassersteinTree}}
\newcommand{\frechetEnergy}{E_F}
\newcommand{\geodesicExtremity}{\mathcal{E}}
\newcommand{\vectorNotation}[1]{\protect\vv{#1}}
\renewcommand{\vectorNotation}[1]{#1}
\newcommand{\axisNotation}[1]{\protect\overleftrightarrow{#1}}
\newcommand{\individualEnergy}{E}
\newcommand{\ensembleSize}{N}
\newcommand{\numberBranchinBarycenter}{N_1}
\newcommand{\numberGeodesicSamples}{N_2}
\newcommand{\planarGridX}{N_x}
\newcommand{\planarGridY}{N_y}
\newcommand{\regularGrid}{G}
\newcommand{\distanceMatrix}{\mathbb{D}}
\newcommand{\maxDimensions}{{d_{max}}}
\newcommand{\projectionOperator}{\mathcal{P}}
\newcommand{\reconstructed}[1]{\widehat{#1}}
\newcommand{\gt}{>}
\newcommand{\lt}{<}
\newcommand{\branch}{b}
\newcommand{\nonLinearFunction}{\sigma}
\newcommand{\batchSequence}{S}
\newcommand{\homologyGroup}{\mathcal{H}}
\newcommand{\bettiNumber}{\beta}
\newcommand{\still}{\mathcal{S}}

\newcommand{\julien}[1]{\textcolor{blue}{#1}}
\renewcommand{\julien}[1]{\textcolor{black}{#1}}

\newcommand{\revision}[1]{\textcolor{red}{#1}}
\renewcommand{\revision}[1]{\textcolor{black}{#1}}

\newcommand{\topoinvis}[1]{\textcolor{blue}{#1}}
\renewcommand{\topoinvis}[1]{\textcolor{black}{#1}}

\newcommand{\josh}[1]{\textcolor{purple}{#1}}
\renewcommand{\josh}[1]{\textcolor{black}{#1}}
\newcommand{\mohamed}[1]{\textcolor{teal}{#1}}
\renewcommand{\mohamed}[1]{\textcolor{black}{#1}}

\newcommand{\minor}[1]{\textcolor{blue}{#1}}

\newcommand{\discuss}[1]{\textcolor{black}{#1}}

\renewcommand{\figureautorefname}{Fig.}
\renewcommand{\sectionautorefname}{Sec.}
\renewcommand{\subsectionautorefname}{Sec.}
\renewcommand{\equationautorefname}{Eq.}
\renewcommand{\tableautorefname}{Tab.}
\newcommand{\algorithmautorefname}{Alg.}
\newcommand{\lineautorefname}{Alg.}

\newcommand{\todo}[1]{\textcolor{red}{#1}}
\renewcommand{\todo}[1]{\textcolor{black}{#1}}

\newcommand{\eqSpace}{\vspace{-0.5ex}}

\newcommand{\mycaption}[1]{
\caption{#1}
}

\newcommand{\myparagraph}[1]{
\noindent\textbf{#1}}

\newcommand{\journal}[1]{\textcolor{blue}{#1}}
\renewcommand{\journal}[1]{\textcolor{black}{#1}}



\firstsection{Introduction}

\maketitle

As computational resources and acquisition devices
are
more efficient and precise,
datasets are growing
in resolution and
details.
This results in a
growth in the size of datasets which
challenges
their
storage, processing and analysis. To address this issue,
\emph{data reduction} is commonly considered,
either
by
employing lossy compression schemes \cite{zfp, tthresh} or by storing reduced
data representations, which concisely, yet precisely, only encode the core
features of the data, possibly during data production (i.e.,
\emph{in-situ} \cite{insitu, AyachitBGOMFM15}).

Topological Data Analysis (TDA) \cite{edelsbrunner09} precisely
focuses on the robust encoding of the structural patterns of a dataset into
concise \emph{topological descriptors}, whose
successful
applications
have been documented in a
variety of domains and contexts \cite{heine16}.
Topological
descriptors
are often used as a mean
of data reduction,
typically by
storing to disk these descriptors instead of the data itself. For instance,
for
time-varying datasets, a typical strategy
\cite{bremer_tvcg11, BrownNPGMZFVGBT21, vestec} consists in storing the actual
data at a low frequency (resulting in the storage of a small number $n$ of
\emph{keyframes}), while storing topological descriptors at a higher frequency
(e.g., for a large number $N \gg n$ of \emph{non-keyframe} time steps). Then,
the produced descriptors can be
directly post-processed by dedicated statistical frameworks
\cite{Turner2014, lacombe2018, YanWMGW20, pont_vis21,
pont_tvcg23, pont_tvcg24}, tailored to the analysis of topological
representations. However, in many scenarios,
a complete investigation might require going back to the original dataset
which initially generated a given topological descriptor, for
instance, for further visual inspection, interpretation and analysis. Then, the
following question arises: \emph{how can we reliably \emph{``invert''} the 
construction of
a topological descriptor?} (i.e., retrieve the dataset which generated
it). Unfortunately, this inverse problem is ill-posed as many distinct datasets
can generate the same topological descriptor
(see \autoref{fig_persistenceDiagrams}). Then, further constraints need to be
considered
\revision{to exploit}
the available data, e.g.,
the
stored
\emph{keyframes}.

\revision{This work addresses this issue by presenting a topology-aware
interpolation approach. Given a reduced representation of an input time-varying
scalar field (i.e., the
persistence diagram of each time-step and a few \emph{keyframes}), the overall
goal of our work is to \emph{invert} the non-keyframe diagrams. To achieve
this goal, we exploit a generative neural architecture to interpolate the
scalar field for a given time  step. This network is trained to
learn the
relation from the time parameter to the input time varying scalar field, based
on the keyframe examples, in order to  reliably extend this relation to the
non-keyframes.}
%
\revision{Also,}
we show how augmenting this architecture with specific
topology-aware losses
\revision{exploiting the input diagrams}
both improves the geometrical and topological reconstruction of the
non-keyframe time steps.
At query time, our approach only
requires as an input the time value for which an interpolation is
desired and it instantaneously produces an interpolated scalar field, via a
single propagation of the time through the neural network.
Experiments interpolating
2D and 3D time-varying ground-truth datasets
demonstrate the superiority of our model,
both in terms of data and topological fitting,
with regard to previous, baseline and neural
interpolation schemes providing comparable query times.

\subsection{Related work}
\label{sec_relatedWork}

This section reviews the literature related to our work, which can be
classified into the following main categories.

\myparagraph{Topological methods in visualization:} The visualization community
has been investigating Topological Data Analysis (TDA) \cite{edelsbrunner09}
for more than two decades \cite{heine16}, with applications to a
variety of
domains, including combustion \cite{bremer_tvcg11}
fluid dynamics \cite{nauleau_ldav22}
material sciences \cite{soler_ldav19},
chemistry \cite{olejniczak_pccp23},
or astrophysics
\cite{shivashankar2016felix} to name a few.
A key feature of TDA is its ability to robustly extract the structural patterns
present in complex datasets and to efficiently represent them into concise
representations. Such representations include persistence
diagrams \cite{guillou_tvcg23},
merge \cite{LukasczykWWWG24} and contour trees
\cite{gueunet_tpds19}, Reeb graphs \cite{gueunet_egpgv19}, or Morse-Smale complexes
\cite{robins_pami11}.
Moreover, another critical aspect of TDA is its  ability to provide
multi-scale hierarchies of the
above topological data representations, enabling in consequence a multi-scale
visualization, exploration and analysis of the topological features of the data.
In that context, \emph{topological persistence} \cite{edelsbrunner02} is an
established importance measure which can be directly read from the persistence
diagram and which can be used to drive the simplification of the above
topological representations. 

In many application scenarios, when
handling time-varying data in scientific computing \cite{bremer_tvcg11,
BrownNPGMZFVGBT21, vestec} (in particular \emph{in-situ} \cite{insitu,
AyachitBGOMFM15}), these topological descriptors are often used as
\emph{proxies} to the data for the purpose of data reduction. For instance, in
the context of simulating mosquito-borne disease spread, Brown et al.
\cite{BrownNPGMZFVGBT21} store time steps of the data at a low frequency to
reduce IO, while the persistence diagram (which is orders of magnitude
smaller than the original data) is stored permanently at a higher frequency.
Similar data reduction strategies based on the merge tree have also been
documented \cite{bremer_tvcg11}.
Then, in a post-process, the resulting ensemble of topological proxies
can be exploited by statistical frameworks
\cite{Turner2014, lacombe2018, YanWMGW20, pont_vis21,
pont_tvcg23, pont_tvcg24}
which are tailored to the analysis of topological
descriptors. In this work, we focus on exploiting these topological data
representations (in particular persistence diagrams), in conjunction with a
sparse temporal sampling of the actual data, to generate plausible
visualizations of the missing, unstored data, in a way that
\revision{favors topological feature preservation.}
For this, we rely on a
neural scheme
which integrates topological constraints, thanks to persistence optimization
\cite{PoulenardSO18, GabrielssonGSG20,
CarriereCGIKU21, SolomonWB21, Nigmetov22, kissi_vis24}, as described in
\autoref{sec_approach}.
\revision{Note that several schemes have been investigated for 
achieving topology-aware compression \cite{soler_pv18, LiLWQYG25}. However,
\topoinvis{to
our knowledge no compression algorithm has been documented for the preservation
of saddle-saddle pairs (a specific type of
features
handled in our work). Also,}
compression is a problem that is \emph{orthogonal}
to the setup studied in our work. First, compressors
have access to
the \emph{full} input data, which eases several aspects dealing with 
\topoinvis{topology enforcement}
and data value preservation. In contrast,
our 
approach does \emph{not} have access to the full input data but only to a 
reduced representation (the persistence diagram of each time step, as well as a 
few \emph{keyframes}). Second, compression could be used in \emph{conjuction} to 
our work, e.g., by using topology-aware compressors to store the
\emph{keyframes}.}


%
%
%
%
%
%

\myparagraph{Neural methods for interpolation:}
The problem of interpolating fields in time appears frequently in both
vision and visualization.

In computer vision, the idea of interpolating video frames from a sequence of 
images is a well known research topic, showing up applications such as producing 
slow-motion videos, temporal upsampling, and video compression.  A recent survey 
\cite{dong2023video}
\revision{categorizes}
approaches into two high level categories: flow-based methods (which 
rely on an estimation of optical flow~\cite{lucas1981iterative}) and 
kernel-based methods (which rely on evaluating differences in a fixed 
neighborhood of each pixel).  This dichotomy is conceptually similar to 
Lagrangian vs. Eulerian methods, as observed by Meyer et 
al.~\cite{meyer2015phase} who also propose looking at phase-based methods for 
video interpolation.

When working in the setting of video, there are
\topoinvis{sharper} differences than one might
consider for field data.  Typically, video footage is assumed to be objects moving 
around in a scene, and thus it is reasonable to use an optical-flow based model that  
tracks the trajectories and velocities of individual pixels (as object samples).  
Techniques to compute optical flow have seen significant advances when using 
deep learning methods such as 
FlowNet~\cite{dosovitskiy2015flownet,ilg2017flownet}, PWC-Net~\cite{sun2018pwc}, 
and RAFT~\cite{teed2020raft}.  While these form an impressive backbone to many 
video interpolation 
techniques~\cite{jiang2018super,bao2019depth,liu2020enhanced,niklaus2020softmax, 
park2020bmbc,park2021asymmetric,huang2022real,reda2022film}, they also can come 
at the cost of complex optimization procedures.  Because of the framing of 
bidirectional flow, often these methods are limited to synthesizing single (or a 
fixed number of) frames between two input frames, and they also make relatively 
strong assumptions (reasonable for videos of objects) that pixel
movement should be captured with flow.  

Kernel-based methods~\cite{lee2020adacof, niklaus2017video,
choi2020channel, kalluri2023flavr}, focus instead 
on
localized 
difference in pixel values within a neighborhood, but as a result often cannot 
estimate motion accurately if between-frame movement is larger than the kernel size.  
The method VideoINR considers interpolating videos in both space and time 
simultaneously~\cite{chen2022videoinr}.  Closely related to our approach is the 
idea of modeling a video as a function in time, as proposed by Chen et al. in 
NeRV~\cite{chen2021nerv}.  We also consider the idea of modeling a time-varying 
field as a network conditioned to produce an entire frame given a time step; 
however, for our work we model the ``frame'' as a 3D volume, and we guide the 
optimization of the network with a topological loss.  Notably, recent work in 
visualization has also built upon NeRV, such as 
NeRVI~\cite{gu2023nervi} and FCNR~\cite{lu2024fcnr}, but these works focus on 
building neural models for rendered images (2D) of time-varying volumes, rather 
than modeling the volume itself.

While most applications in computer vision are limited to 2D + time, some of
these methods have been extended to 3D + time images, particularly coming from
medical imaging~\cite{guo2020spatiotemporal,wang2021sparse}.  Yin et
al. considered adapting some of the above methods directly to the setting of
coronary angiography~\cite{yin2021analysis}.
More generally, to address modeling volumetric data, in visualization numerous
authors have begun to consider machine learning methods 
for
time-varying volumes.  TSR-TVD is 
recent
work in this space utilizing recurrent 
generative models for temporal super resolution~\cite{han2019tsr}.  Han et al. 
later considered the generalized problem of spatio-temporal super-resolution in 
STNet~\cite{han2021stnet}.  Both  methods suffer from long training
times and only produce a fixed temporal scale factor, suffering from the
same limitations as many video interpolation methods.  STSRNet employs
\topoinvis{a}
two-stage framework using optical flow for spatial-temporal super 
resolution~\cite{an2021stsrnet}.  Recently, FLINT~\cite{gadirov2024flint} 
also used optical flow 
for temporal super resolution, building 
upon RIFE~\cite{huang2022real}.

Following the introduction of implicit neural representation (INRs) to the 
visualization community for volume compression by Lu et 
al.~\cite{lu2021compressive}, numerous authors considered their use in other 
applications.  CoordNet leveraged INRs for multiple visualization tasks 
(including spatio-temporal super resolution and visualization synthesis) for 
time-varying ensembles~\cite{han2022coordnet}.  While this approach produced a 
generalized framework, to address large training times and model sizes, 
different authors considered using knowledge distillation for either learning 
hypernetworks~\cite{wu2023hyperinr} or compressed models~\cite{han2023kd} with 
CoordNet serving as the teacher model.  FFEINR~\cite{jiao2024ffeinr} considered 
an INR for joint spatio-temporal super resolution based on 
VideoINR~\cite{chen2022videoinr}.  Finally, STSR-INR is a recent 
neural method for 
spatio-temporal super resolution, supporting multivariate ensemble 
data~\cite{tang2024stsr}.  We consider STSR-INR as a representative example of
state-of-the-art INRs to compare \topoinvis{against}, although it differs from
our work in
that it does not explicitly consider a loss based on topological features.
Including such a loss directly into a coordinate-based network (that predicts
positional samples of the field in batches, rather than an entire volume) would
require a significantly more costly loss function.

\subsection{Contributions}
\label{sec_contributions}

\begin{figure*}
\centering
\includegraphics[width=\linewidth]{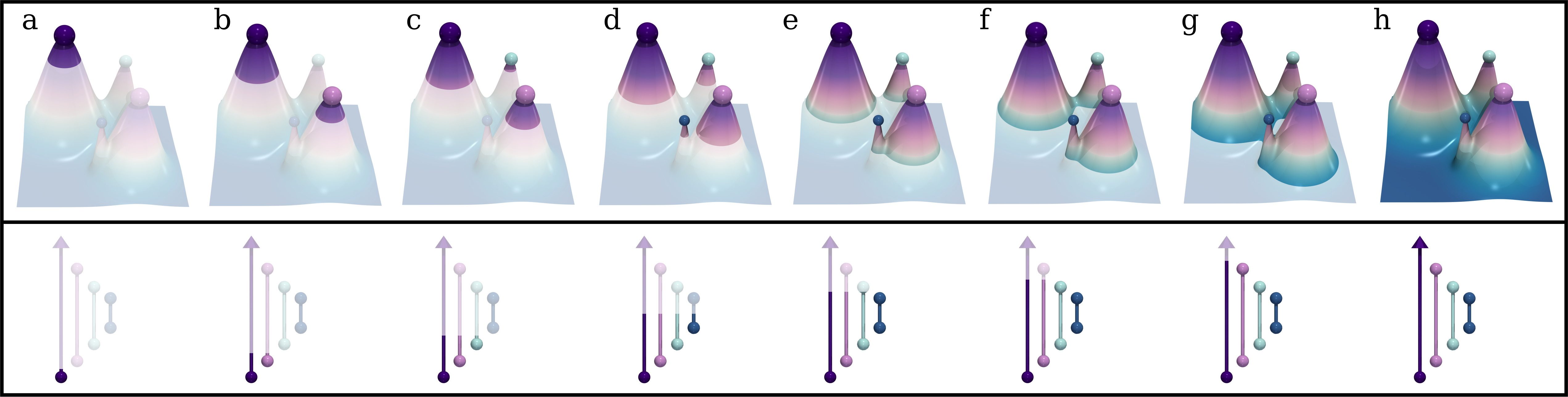}
\caption{\revision{Filtration of the opposite elevation function (going
downward, purple: low values, cyan: high values) on a toy
terrain example (in transparent). From left
to right, simplices are progressively added in the filtered simplicial
complex for increasing values of opposite elevation. Each local minimum
triggers the \emph{birth} of a connected component in the complex
(sub-figures \emph{a} to \emph{d}). Connected components are represented by
growing bars of matching color in the diagram (bottom).
When a component merges with an \emph{older} one (sub-figures \emph{e},
\emph{f} and \emph{g}), its corresponding bar terminates its growth in the
diagram and the corresponding topological feature is said to \emph{die} at the
corresponding
value.
}}
\label{fig:filtration}
\end{figure*}

This paper makes the following new contributions:
\begin{enumerate}[itemsep=-.75mm]
  \item \revision{An approach combining  \emph{(i)} a generative neural model 
for scalar field interpolation with \emph{(ii)} \josh{topological losses based
on} persistence 
diagrams, for constraining the topology \emph{and} geometry of the output
interpolations.}


  \item
  \revision{A TTK/PyTorch}
  implementation
  for reproducibility.
\end{enumerate}


\section{Background}
\label{sec_background}

This section provides the required technical background in Topological Data
Analysis (TDA). For further readings, we refer the reader to textbooks on
computational topology \cite{edelsbrunner09}.

\subsection{Input data}
\label{sec_inputData}

Each input dataset is typically provided as a scalar field $f :\cubicalComplex
\rightarrow
\mathbb{R}$, valued on the vertices of a cubical complex $\cubicalComplex$
(i.e., a 2D or 3D regular grid).
In practice, for implementation
genericity purposes, $\cubicalComplex$ is often triangulated
(via an on-the-fly emulation \cite{ttk17}) into a simplicial complex $\domain$
(typically, with the Freudenthal triangulation).
This yields a piecewise linear (PL) scalar field $\rho : \domain \rightarrow
\mathbb{R}$, valued on the vertices of $\domain$ and linearly interpolated on
the simplices of higher dimensions.
In practice, the data values are
given as
a
\emph{data vector}, noted $\dataVector_f \in \mathbb{R}^{\numberOfVertices}$
\topoinvis{(where $\numberOfVertices$ is the number of vertices in $\domain$)}.
Additionally, $f$ is required to be injective on the vertices,
which
is achieved
via a variant of simulation of simplicity \cite{edelsbrunner90}.
\revision{Also,}
in our experiments, $f$ will be normalized (i.e., valued between $0$
and $1$) to ease parameter tuning across datasets.

\begin{figure}
\centering
\includegraphics[width=\linewidth]{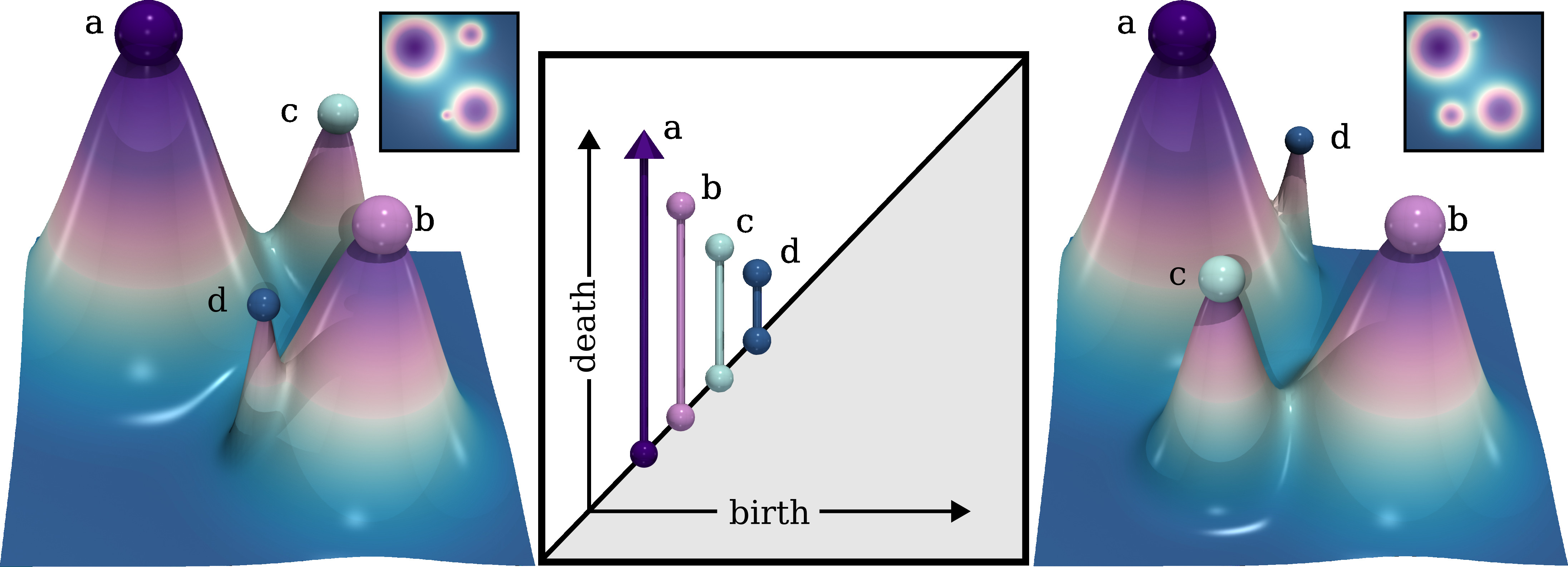}
\caption{Scalar fields (height opposite) admitting a common
persistence diagram (center). Each hill is encoded in the diagram by a vertical
bar whose length encodes the \emph{persistence} of the corresponding topological
feature in the data (arrows indicate generators with infinite persistence).
While the diagram encodes the list of topological features
with their birth, death and persistence, it forgets their
geometrical
realization in the data.}
\label{fig_persistenceDiagrams}
\end{figure}

\subsection{Persistence diagrams}
\label{sec_persistenceDiagrams}

Persistent homology emerged independently from multiple research groups
\cite{B94, frosini99, robins99, edelsbrunner02}. Conceptually, it analyzes a
progressive sweep of the data (called \emph{filtration}),
and computes at each stage its
associated topological features (specifically, its \emph{homology
generators}). Additionally, it establishes correspondences
between features detected at successive steps. This allows for the
identification of topological features along with their duration throughout the
filtration process.

In our work,  we rely on the lexicographic filtration of the input scalar
data, which sorts simplices by extending the global order on the vertices (based
on their $f$ values) to the simplices (see \cite{guillou_tvcg23}). This induces
a nested
sequence of simplicial complexes $\emptyset =  \domain_0 \subset \domain_1
\subset \dots \subset \domain_{\numberOfSimplices} = \domain$ (with
$\numberOfSimplices$ being the number of simplices in $\domain$).

At each stage $i$ of the filtration, for a given $p^{th}$ homology group, one
can track the corresponding \emph{homology generators}, whose number defines the
$p^{th}$ Betti number $\bettiNumber_p$.
In low dimensions (e.g., 3D), this quantity is accompanied with a simple
intuition: $\bettiNumber_0$ encodes the number of connected components,
$\bettiNumber_1$ that of topological handles and $\bettiNumber_2$ that of
cavities in $\domain_i$.
Given two consecutive steps $i$ and $j$ of the filtration,
since the corresponding simplicial complexes $\domain_i$ and
$\domain_j$
are nested,
one can easily establish a correspondence (\revision{i.e.,} a 
\emph{\revision{homomorphism}})
between the generators observed at step $i$ and those observed at step $j$. In
particular, a \emph{persistent generator} is said to be born at step $j$ if it
has not been matched to any  generator pre-existing at step $i$. By symmetry, a
generator is said to die at step $j$ if it merges with another older homology
class (born before it). Then, each $p$-dimensional persistent generator is
accompanied by
a pair of simplices $(\simplex_b, \simplex_d)$ called
\emph{persistence pair}, which corresponds to the $p$ and $(p+1)$ simplices
inserted in the filtration at the birth and death of the generator. The
\emph{persistence} of the pair is given by $p(\simplex_b,
\simplex_d) = \topoinvis{\dataVector_f[v_d] - \dataVector_f[v_b]}$, where $v_b$
and $v_b$
are the \emph{birth} and \emph{death} vertices of the pair (i.e., the highest
vertices in $\simplex_b$ and $\simplex_d$). Note that the
generators characterizing the homology groups of the input complex
$\domain$ are said to have \emph{infinite persistence}.
\revision{This process is illustrated in \autoref{fig:filtration} for the 
specific case of the $0$-dimensional persistent homology
(representing
persistent connected components).}

The \emph{persistence diagram} is a concise encoding of the persistence pairs
of a scalar field (\autoref{fig:filtration}). It represents each pair
by a vertical bar, whose bottom point is placed on the diagonal and whose top
point is placed at coordinate
\topoinvis{$\bigl(\dataVector_f[v_b],
\dataVector_f[v_d]\bigr)$}. Bars representing generators with
infinite persistence are cropped at the maximum $f$ value.
Note that, as discussed in the introduction, the pre-image of the construction
of a persistence diagram is not unique. This is shown in
\autoref{fig_persistenceDiagrams}, where two distinct scalar fields produce the
same diagram. This further illustrates the ill-posedness of inverting the
construction of persistence diagrams.

\subsection{Wasserstein distance between diagrams}
\label{sec_wasserstein}

\begin{figure}
\includegraphics[width=\linewidth]{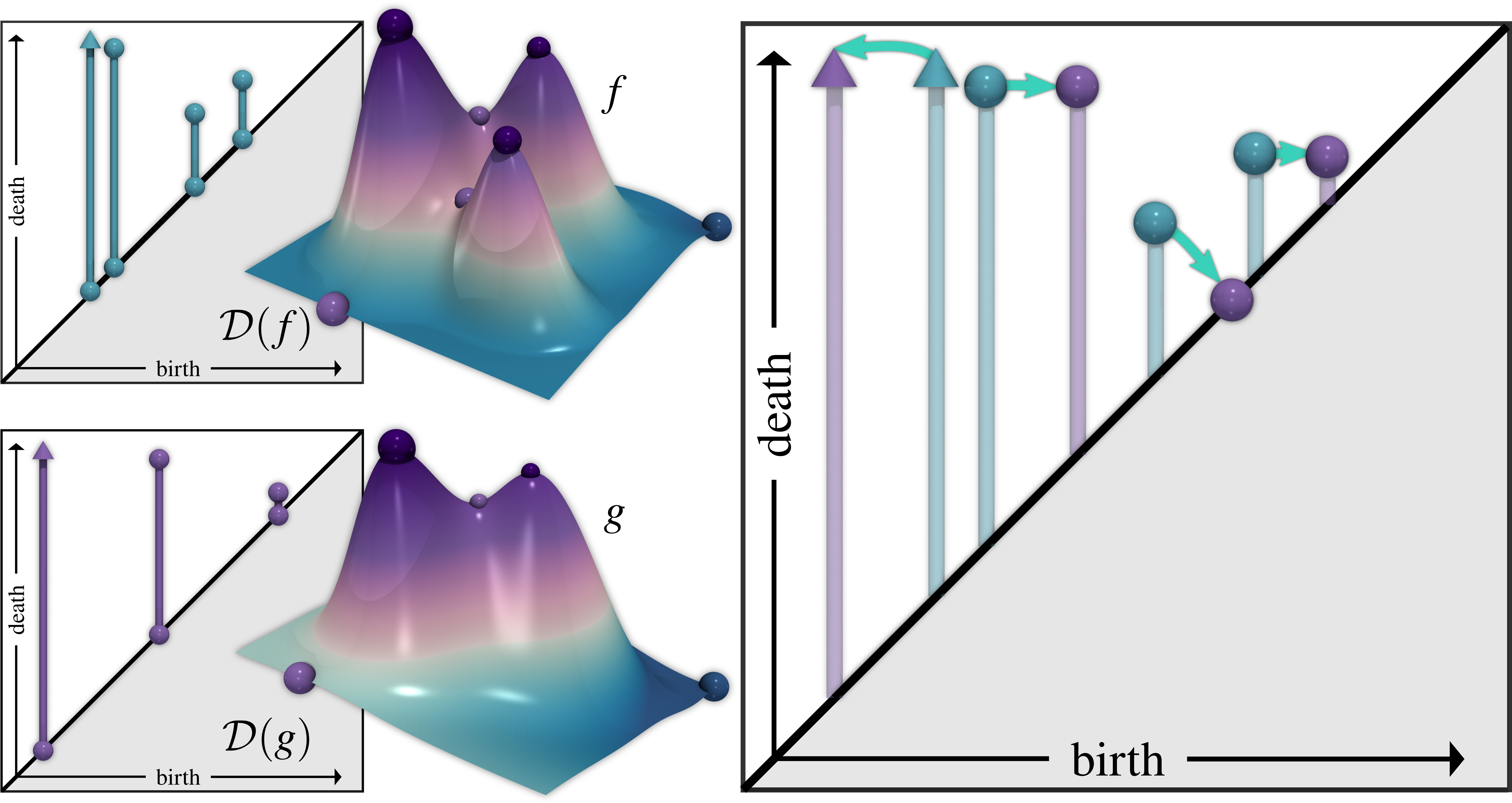}
  \caption{The Wasserstein distance $\wasserstein{2}$ between $\diagram(f)$
(top) and $\diagram(g)$ (bottom) is
obtained
by assignment
optimization (\autoref{eq_wasserstein}) in the birth-death plane
(right).
The optimal assignment $\phi^*$  (arrows)
defines
a least-effort deformation
of $\diagram(f)$ into $\diagram(g)$, by moving
persistence pairs in the plane.
}
  \label{fig_wassersteinDistance}
\end{figure}

The \emph{Wasserstein distance} (\autoref{fig_wassersteinDistance}) is an
established metric for comparing
two diagrams $\diagram(f)$ and $\diagram(g)$ \topoinvis{\cite{edelsbrunner09}}.
To facilitate its computation,
the input diagrams first need to be \emph{augmented}, to guarantee
that they count the same number of points. This process is achieved as follows.
Given a point $p = (p_b, p_d) \in \diagram(f)$, let
$\projection(p)$ be its diagonal projection: $\projection(p) =
\big({{1}\over{2}}(p_b + p_d), {{1}\over{2}}(p_b
+ p_d)\big)$. We note $\projection_f$ and $\projection_g$ the sets of the
diagonal projections of the points of $\diagram(f)$ and
$\diagram(g)$. Then, $\diagram(f)$ and $\diagram(g)$ are
\emph{augmented} by
adding
to each of them the set of diagonal points $\projection_g$
and $\projection_f$ respectively. After this step, we have
$|\diagram(f)| = |\diagram(g)|$.

Then, given two augmented persistence diagrams $\diagram(f)$ and $\diagram(g)$,
the $L^q$ Wasserstein distance between them is defined as:
\begin{eqnarray}
\label{eq_wasserstein}
\wasserstein{q}\big(\diagram(f), \diagram(g)\big) = \min_{\phi \in \Phi}
\Big(\sum_{p \in \diagram(f)} c\big(p, \phi(p)\big)^q\Big)^{{{1}\over{q}}},
\end{eqnarray}

\noindent
where $\Phi$ is the set of bijections from
$\diagram(f)$ to $\diagram(g)$,
such that points of finite
(respectively infinite) $r$-dimensional persistent generators are mapped to
points of
finite (respectively infinite) $r$-dimensional persistent generators.
%
In particular,
the cost $c(p, p')$ is set to $0$ when both $p$ and $p'$
are diagonal points (i.e., dummy features).
Otherwise, it is set to the Euclidean distance in the birth-death plane
$||p - p'||_2$.

\topoinvis{To analyze our experimental results (\autoref{sec_results}),
in particular to compare topology preservation
across multiple datasets, we will consider the following normalization:
\begin{eqnarray}
\nonumber
  \normalizedWasserstein{q}\big(\diagram(f), \diagram(g)\big) =
  {{\wasserstein{q}\big(\diagram(f), \diagram(g)\big)}\over{
  \Big(
    \wasserstein{q}\big(\diagram(f), \diagram_{\emptyset}\big)^q
    + \wasserstein{q}\big(\diagram_{\emptyset}, \diagram(g)\big)^q
  \Big)^{{{1}\over{q}}}
  }},
\end{eqnarray}
where $\diagram_{\emptyset}$ is the empty diagram. This quantity is $0$ when
$\diagram(f)$ and $\diagram(g)$ are identical and $1$ when there is no matching
between the features of the diagrams (i.e., when the optimal assignment
only sends points to/from the diagonal).}
%

\begin{figure*}
  \centering
      \centering
      \includegraphics[width=\linewidth]{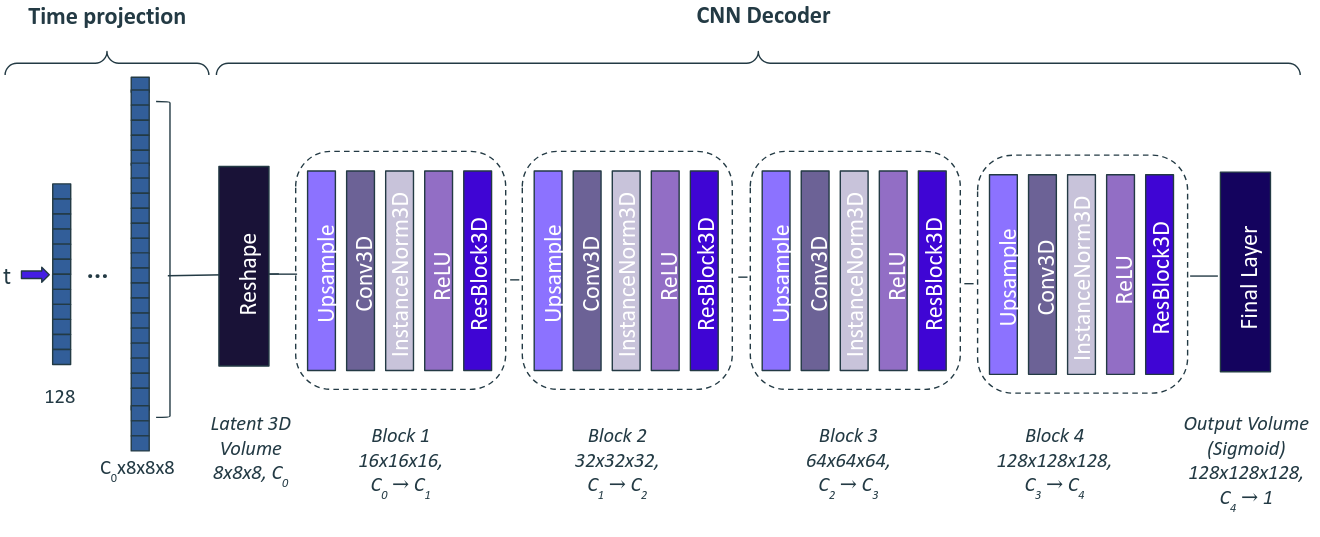}
      \caption{%
      \emph{TimeToScalarField}:
      An input value $t$
is encoded using sinusoidal positional encoding and projected through a fully
connected
layer into a latent 3D tensor of size $C_0\times8\times8\times8$, where $C_0$
denotes
the initial number of channels.
This tensor is processed by a CNN
composed of four sequential blocks,
each consisting
of: \emph{(i)} trilinear upsampling (by a factor of 2),
\emph{(ii)} 3D convolution (with kernel size $3\times3\times3$),
\emph{(iii)} instance normalization,
\emph{(iv)} ReLU activation, and
\emph{(v)} a residual block
(\emph{ResBlock3D}, \cite{he2016deep}). The spatial
resolution is
progressively increased while the number of channels is reduced. A final
convolution followed
by a Sigmoid activation produces the output volume (here, with resolution
$128^3$).
}
  \label{fig_architecture}
\end{figure*}

\subsection{Persistence optimization}
\label{sec_persistenceOptimization}

Persistence optimization has been investigated by several authors
\cite{PoulenardSO18, GabrielssonGSG20,
CarriereCGIKU21, SolomonWB21, Nigmetov22}.
\topoinvis{We}
recap here an
efficient and generic framework
\cite{CarriereCGIKU21}, which has been recently exploited
in visualization  \cite{kissi_vis24} \topoinvis{and which is used by our
work
to enforce topological constraints.}



The goal of persistence optimization is to modify an input data vector
$\dataVector_f \in \mathbb{R}^{\numberOfVertices}$
(\autoref{sec_inputData}), such that its persistence diagram $\diagram(f)$
minimizes the considered loss
$\loss$. For that, $\loss$ is typically decomposed
into several steps. First, we call the \emph{filtration map} the function
$\filtration : \mathbb{R}^{\numberOfVertices} \rightarrow
\mathbb{R}^{\numberOfSimplices}$, which maps a
data vector $\dataVector_f$
to a filtration,
encoded
as a vector
$\filtration(\dataVector_f) \in \mathbb{R}^{\numberOfSimplices}$,
which stores in its $i^{th}$  entry the global lexicographic order
(\autoref{sec_persistenceDiagrams}) of the $i^{th}$ simplex
of
$\domain$.
In practice, a \emph{backward filtration map}, noted $\filtration^{+} :
\mathbb{R}^{\numberOfSimplices} \rightarrow
\mathbb{R}^{\numberOfSimplices}$, is also maintained. It maps a filtration
vector
$\filtration(\dataVector_f)$ back to a vector in
$\mathbb{R}^{\numberOfSimplices}$,
whose $i^{th}$ entry is the index of the highest vertex (in global vertex
order) of the $i^{th}$ simplex in the global lexicographic order.
Next, a \emph{persistence map} $\persistenceMap:
\mathbb{R}^{\numberOfSimplices} \rightarrow \mathbb{R}^{\numberOfSimplices}$
maps a filtration vector $\filtration(\dataVector_f)$ to a
\topoinvis{vector}
$\persistenceMap\big(\filtration(\dataVector_f)\big)$. In particular, the
$i^{th}$ entry of this map assigns the $i^{th}$ simplex in the global
lexicographic order to its \emph{critical simplex persistence order}. That
latter order is obtained as follows.
%
First,
the points of
$\diagram(f)$ are sorted by \emph{diagram order}, i.e.,
by increasing birth and then by
increasing death (to disambiguate ties).
The
persistence pairs
are then sorted
with
this
order, by interleaving the birth and death simplices of  each
diagram point. This provides an ordering of the critical simplices, called
the
\emph{critical simplex
persistence order}, where the $(2i)^{th}$ and $(2i+1)^{th}$ entries encode
respectively  the birth and death simplices of the $i^{th}$ point $p_i$ in
the diagram order.
Critical simplices corresponding to homology classes of
infinite persistence are added at the end of this list,
in increasing order of
$f$
values.
Finally, in the persistence map, the entries corresponding
to regular simplices
are set to $-1$.


Now, let us consider a generic energy term $\energy:
\mathbb{R}^{\numberOfSimplices} \rightarrow
\mathbb{R}$, which given a persistence
\topoinvis{map}
(defined above), evaluates the
relevance of the corresponding diagram $\diagram(f)$ for the
considered problem.
Then, given an input
vector
$\dataVector_f$, the associated loss $\loss : \mathbb{R}^{\numberOfVertices}
\rightarrow \mathbb{R}$ is:
\begin{eqnarray}
\nonumber
\loss(\dataVector_f) = \energy \circ \persistenceMap \circ
\filtration(\dataVector_f).
\end{eqnarray}

As discussed in  \cite{CarriereCGIKU21}, if
$\energy$ is locally Lipschitz
and a definable function of persistence, then we have the guarantee that
the composition $\energy \circ \persistenceMap \circ \filtration$ is also
definable and locally Lipschitz \cite{CarriereCGIKU21}, which itself guarantees
%
that
the generic loss
$\loss$ is differentiable almost everywhere.
Then, stochastic gradient descent \cite{KingmaB14} can be employed with
guaranteed convergence \cite{DavisDKL20}.
Assuming a constant
global
lexicographic order,
the loss is
minimized
by displacing
each point $p_i$ in
the diagram $\diagram(f)$ according to the gradient.
Concretely, this displacement
is
back-propagated
in the data
vector $\dataVector_f$, by identifying the vertices $v_{i_b}$ and $v_{i_d}$
corresponding to the birth and death
of the
$i^{th}$ point in the diagram order:
\begin{equation}
\begin{array}{r@{}l}
\label{eq_diagramToVertex}
\nonumber
v_{i_b} &{}= \filtration^{+}\big(\persistenceMap^{-1}(2i)\big)\\
v_{i_d} &{}= \filtration^{+}\big(\persistenceMap^{-1}(2i + 1)\big),
\end{array}
\end{equation}
and by updating their data values $\dataVector_f(v_{i_b})$ and
$\dataVector_f(v_{i_d})$ according
to the gradient.
This back-propagation mechanism (updating
the input vector $\dataVector_f$ to decrease the loss $\loss$) ensures that
persistence optimization can be used in
conjunction with the automatic differentiation features of
neural networks \cite{pytorch}, thereby enabling their training via gradient
descent \cite{KingmaB14}, possibly in conjunction to
extra
regularization
terms, as described next.

\section{Approach}
\label{sec_approach}

This section presents our \revision{overall interpolation technique},
given an input time-varying sequence of persistence diagrams, along with a
sparse temporal sampling of the corresponding scalar fields.
\revision{While the neural network architecture exploited in our \josh{approach} is
typical of 
related work \cite{chen2021nerv}, we still document its specifications in 
\autoref{sec_architecture} for completeness and reproducibility purposes.}

\subsection{Overview}
Our approach relies on a generative neural network architecture, 
\revision{referred to in the remainder as}
\emph{TimeToScalarField}. It is  presented in \autoref{sec_architecture} and
schematically illustrated in \autoref{fig_architecture}. It takes as an input a
time parameter value $t \in [0, 1]$, and generates a scalar data vector
$v_{f(t)} \in \range^{n_v}$
defining  a scalar field $f(t)$ on a
cubical complex $\cubicalComplex$ (i.e., a regular grid) with dimensions $c_x,
c_y, c_z$ assumed to be multiples of powers of two (i.e., $n_v = c_x \times
c_y \times c_z$).

The training of this architecture is performed on the $N$ time-steps of the
input temporal sequence, 
specifically with:
\vspace{-.75ex}
\begin{itemize}[itemsep=-.75mm]
\item the $n \ll N$ keyframe scalar fields and
persistence diagrams,
\item
\topoinvis{and}
the
 $(N-n)$ non-keyframe persistence diagrams (for which the scalar fields are
\emph{not} given).
\end{itemize}
\vspace{-.75ex}
Note that the interpolation of distinct input temporal sequences requires the
training of distinct networks.
The training is achieved
by minimizing the overall loss presented in \autoref{sec_losses}.
In particular, this optimization is carried out by stochastic gradient descent
\cite{KingmaB14}, relative to the gradient of the training loss, obtained via 
automatic
differentiation \cite{pytorch}.
Note that for the $(N-n)$ non-keyframes, the
corresponding scalar data is not available (only the persistence diagrams are
available). Therefore, the loss terms involving scalar fields  (e.g.,
\emph{Mean Squared Error}, MSE) are set to zero for non-keyframe time values
(see \autoref{sec_losses}).
At query time, the time $t \in [0, 1]$ for which an interpolation is
desired is presented to the input of the trained neural network, which
propagates it to
return the
vector $v_{f(t)}$, defining the output scalar field
$f(t) : \cubicalComplex \rightarrow \range$.

%

\subsection{Architecture}
\label{sec_architecture}

\revision{The}
\emph{TimeToScalarField}
architecture (\autoref{fig_architecture}) is
designed to
generate spatially structured scalar fields from time values.
It includes
\emph{(i)} a time projection
and
\emph{(ii)} a
\emph{convolutional neural network} (CNN)\topoinvis{.}

\myparagraph{Time projection:} The input time value $t \in [0, 1]$ is first
presented to a \emph{positional encoding} (PE) layer \cite{gehring17,
vaswani2017attention} as it is often reported
(and confirmed by our experiments)
to improve temporal coherence. This layer is \emph{fixed} (i.e., not
optimized) and simply maps $t$ to a high-dimensional vector
$PE(t) \in \mathbb{R}^{T}$.
Each entry $i$ of this vector is
a sinusoidal function of $t$, with increasing frequencies for increasing values
of $i$ \cite{vaswani2017attention}:
\begin{eqnarray}
\nonumber
\begin{cases}
    \bigl[PE(t)\bigr]_{2i} = \sin(\julien{2\pi} t / 10,000^{2i/T})\\
    \bigl[PE(t)\bigr]_{2i+1} = \cos(\julien{2\pi}t / 10,000^{2i/T})
  \end{cases}.
\end{eqnarray}
This first positional encoding $PE(t)$ is then projected into a higher
dimensional latent space $\mathbb{R}^{C_0 \times r_0 \times r_0 \times r_0}$
\topoinvis{(where $r_0$ is a coarse, initial data resolution for the
subsequent CNN decoder, see below)}
via
sequential, fully connected layers
with ReLU activations 
(\emph{``Reshape''} layer, 
\autoref{fig_architecture}). This projection enriches the latent
representation and provides a robust foundation for the subsequent decoding.
Such a re-projection approach has been widely validated in
conditional generative frameworks \cite{isola2017image}.
The output of that layer forms the input for the subsequent CNN decoder.


\myparagraph{CNN decoder:} Our CNN decoder (\autoref{fig_architecture}, right)
reconstructs the output data progressively through successive upsampling stages.
The decoder starts with a coarse resolution
($r_0\times r_0\times r_0$ in 3D and $r_0\times r_0$ in 2D)
over $C_0$ \emph{channels} (i.e., $C_0$ instances of the optimization,
started at distinct random initializations).
As recommended in the CNN
literature, $C_0$ is progressively decreased 
via channel-wise filtering
at each stage $i$
into $C_i$, until reaching $1$ (see \autoref{sec_datasets} for further 
discussions).
The initial resolution $r_0$ is multiplied by $2$ at each stage $i$ along each 
dimension until reaching the input
resolution (i.e., $c_x \times c_y \times c_z$). 
Specifically, each processing block $i$
(dashed boxes in \autoref{fig_architecture}) consists of:
\vspace{-.75ex}
\begin{enumerate}[itemsep=-.75mm]
\item
an upsampling phase (with bilinear and trilinear interpolants in 2D 
and 3D respectively),
\item
a convolutional block (with kernel sizes $3 \times 3$ and $3 \times 
3 \times 3$ in 2D and 3D respectively),
\item
instance normalization \cite{ulyanov2016instance} (to
reduce overfitting and improve generalization)
\item
a non-linear
activation (ReLU)
and 
\item
a residual block \cite{he2016deep}
\revision{(to stabilize}
the training and
mitigate
vanishing gradient issues often encountered in deep CNNs).
\end{enumerate}
\vspace{-.75ex}



%
%

%
%
%
%
%
%
%
%

\subsection{Losses}
\label{sec_losses}

Given a batch of input values $t$,
the output
predictions
provided by the
neural network are evaluated with the following overall loss:
\begin{eqnarray}
\nonumber
\loss = \loss_{MSE} + \alpha \loss_{\nabla}
  + \beta \loss_{CV} + \gamma \loss_{\wasserstein{2}}.
\end{eqnarray}
It is composed of four terms, detailed below.


\myparagraph{Data fitting:} This term is the traditional \emph{Mean Squared
Error} (MSE) which evaluates, only for the $n$ keyframes (it is
$0$ otherwise), the fitting
of each prediction $\topoinvis{f_{(t_k)}}$ to its training keyframe
$f_k$:
\begin{eqnarray}
\nonumber
\loss_{MSE} = {{1}\over{n}} \sum_{k = 1}^{n}
\todo{{{1}\over{n_v}} \sum_{j = 1}^{n_v} }||
\topoinvis{f_{(t_k)}(v_j)}
- f_k(v_j) ||^2_2.
\end{eqnarray}

\myparagraph{Gradient fitting:} To improve the
\topoinvis{geometry}
of fine scale details, an additional term is
considered, only for the $n$ keyframe timesteps (it is set to $0$ otherwise), 
to evaluate the fitting of the
gradient of each network prediction $\topoinvis{f_{(t_k)}}$ to that of its
training
keyframe
$f_k$:
\begin{eqnarray}
\nonumber
\loss_{\nabla} = {{1}\over{n}} \sum_{k = 1}^{n}
\todo{{{1}\over{n_v}}}
\sum_{j = 1}^{n_v} \sum_{i =
1}^{n_d} \loss_{\nabla_1} \Bigl(\bigl(\nabla
\topoinvis{f_{(t_k)}(v_j)}\bigr)_i,
\bigl(\nabla f_k (v_j)\bigr)_i\Bigr),
\end{eqnarray}
where $n_d$ is the dimensionality of the dataset (in our experiments, $2$ or 
$3$), where $\nabla$ is the vector formed by the partial derivatives of the 
scalar
field at a vertex
$v_j$ in the geometrical domain $\cubicalComplex$, and where $\loss_{\nabla_1}$ 
is the so-called \emph{smooth $L_1$ loss} \cite{Girshick15} (which presents the 
practical interest of the $L_1$ norm, while still being differentiable in $0$):
\begin{eqnarray}
\nonumber
  \loss_{\nabla_1}(x, y) = 
\begin{cases} 
0.5 (x - y)^2, & \text{if } |x - y| < 1\text{,}\\
|x - y| - 0.5, & \text{otherwise.}
\end{cases}
\end{eqnarray}

\myparagraph{Critical values:} For each of its bars, in addition to its birth
and death values,
the persistence diagram also typically encodes in practice the
identifiers $v_b$ and $v_d$ of the vertices respectively implied in the birth
and death of the corresponding topological feature
(\autoref{sec_persistenceDiagrams}).
The set of all birth and death vertices form the
\emph{critical points} of the underlying piecewise linear
scalar field
\topoinvis{\cite{edelsbrunner09}.}
Specifically, the scalar value of a birth (respectively death) vertex $v_b$
(respectively $v_d$) is given by the birth (respectively death) of its bar in
the diagram.
This information, which is
available for all the $N$ timesteps, can be re-used to enforce the scalar value
at the precise location of each critical point, helping preserve the
location of the topological features in the
geometrical
domain \topoinvis{(see \autoref{sec_influence}).}
This can be achieved with the following
loss, which
evaluates,
for each prediction $\topoinvis{f_{(t_k)}}$, the fitting between the pointwise
value $\topoinvis{f_{(t_k)}(v_j)}$ and the corresponding
value $f_k(v_j)$ given
by the input persistence diagram
$\diagram_k$
for each vertex $v_j$ in the set of critical
points $CP_k$ of $f_k$:
\begin{eqnarray}
\nonumber
\loss_{CV} = {{1}\over{N}} \sum_{k = 1}^{N} {{1}\over{n_{CP_k}}}
\sum_{j =
1}^{n_{\topoinvis{CP}_k}} || \topoinvis{f_{(t_k)}(v_j)} - \topoinvis{f_k(v_j)}
||^2_2.
\end{eqnarray}

\myparagraph{Topology correction:}
\revision{To} enforce topological
\revision{preservation,}
we evaluate the
topological fitting
between the diagrams $\diagram(v_{f(t_k)})$ and their target inputs
$\diagram_{k}$ with the Wasserstein distance (\autoref{sec_wasserstein}):
\begin{eqnarray}
\nonumber
\loss_{\wasserstein{2}} = {{1}\over{N}} \sum_{k = 1}^{N}
\wasserstein{2}\bigl(\diagram(\topoinvis{f_{(t_k)}}), \diagram_{k}\bigr).
\end{eqnarray}
\revision{This loss effectively quantifies and penalizes topological errors.}
Since the Wasserstein distance is a definable function of persistence
\cite{CarriereCGIKU21},
it can be used within
the
optimization framework
from
\autoref{sec_persistenceOptimization} (i.e., $\mathcal{E} =
\loss_{\wasserstein{2}}$),
with guaranteed convergence
\cite{DavisDKL20, CarriereCGIKU21},
\topoinvis{in conjunction with the
above (convex and \julien{differentiable}) losses.}


\subsection{Computational details}
\label{sec_details}

This section provides practical details for the training of the model
presented in \autoref{sec_architecture} with regard to the
overall
loss described in
\autoref{sec_losses}.
The training is organized into two phases:
\vspace{-.75ex}
\begin{enumerate}[itemsep=-.75mm]
\item
\emph{scalar field training}
(with the following loss weights, \autoref{sec_losses}: $\alpha = 0.1$, $\beta =
1$ and $\gamma = 0$)
and

\item
\emph{topology correction} (with loss weights: $\alpha = 0$,
$\beta=1$, $\gamma = 1$). 
\end{enumerate}
\vspace{-.75ex}
\revision{This two-phase strategy is motivated by the computational effort 
required by the \emph{topology correction} step (which involves the 
computation of a persistence diagram at each iteration). Then, this strategy 
first learns quickly a plausible geometry for the missing scalar fields 
(\emph{scalar field training}) prior to optimizing their topology with more 
expensive computational efforts (\emph{topology correction}). Also, from a 
practical standpoint, our initial experiments reported that this two-phase 
strategy improved the convergence of the training.}

For each phase, the entire input ($N$ diagrams, $n$
keyframes) is presented to the network at each epoch, for a number $n_1$ and 
$n_2$ of epochs for the two phases respectively (see \autoref{sec_datasets} for 
further 
discussions).
The purpose of this decomposition is to first generate a good estimation of the
 scalar fields in phase 1, prior to refining their topology in
phase 2 (which is significantly more  expensive computationally). Specifically,
in phase 2, the last layer of temporal projection as well as the
first block of the CNN decoder are \emph{frozen}, in order to maintain through
phase 2 the large scale details learned in phase 1. Also, to mitigate
overfitting and improve generalization, dropout regularization \cite{wan2013}
is used, omitting randomly the update of a network weight,
with a probability set to $0.1$.

Each epoch of topology correction (phase 2) requires the computation of $N$
persistence diargams $\diagram(v_{f(t_j)})$
as well as the
estimation of the Wasserstein distances
to their
input targets $\diagram_{j}$, which is done in practice in
quadratic time (with the number of vertices $n_v$ in the output grid).
To
accelerate this process, we \emph{prune} each input target diagram
$\diagram_{j}$ by removing its bars with a persistence
(\autoref{sec_persistenceDiagrams}) smaller than $1\%$ of its largest bar, which
is a typical persistence thresholding in the applications. We prune
similarly $\diagram(v_{f(t_j)})$ and fix the assignment of the pruned bars to
the diagonal (which is equivalent to the destruction of these noisy features).
This two-stage pruning reduces the size of the diagrams and drastically
accelerates the optimal assignment optimization at the basis of the Wasserstein
distance computation (\autoref{sec_wasserstein}). Finally, we use shared-memory
parallelism to compute each diagram (and its distance to its target) in a
distinct task.

The minimization of the overall loss (\autoref{sec_losses}) is performed with
the Adam solver \cite{KingmaB14} (with weight decay set at $10^{-6}$, to
mitigate overfitting), with a relatively low initial learning rate 
to favor a stable optimization (see \autoref{sec_datasets} for numerical 
values).

\topoinvis{Finally,
this
architecture (\autoref{sec_architecture}) is subject to
additional
meta-parameters
e.g., $T = 128$, $r_0 =
8$.
Overall, all the meta-parameters of our approach were adjusted
empirically to optimize performance
(see \autoref{sec_datasets}, for dataset specific parameters).}







\section{Results}
\label{sec_results}

This section presents
results obtained with a PyTorch
\cite{pytorch} implementation of our
approach,
using TTK \cite{ttk17, ttk19}
for persistence
computation and matching.
Experiments were
performed
in a Google Colab
environment,
with an Nvidia A100-SXM4 GPU (RAM: 40 GB) and an Intel Xeon CPU
(2.2 GHz, 6
cores,
RAM: 80 GB).

\subsection{Test datasets}
\label{sec_datasets}
Our experiments include both synthetic and real-life time-varying
2D and 3D scalar fields.
For convenience, the considered 2D (respectively 3D) datasets have all been
resampled to an initial resolution of $512^{2}$ (respectively
\todo{$128^{3}$}). Moreover, to replicate a setting that is typical of in-situ
data production \cite{bremer_tvcg11, BrownNPGMZFVGBT21, vestec}, we selected as
\emph{keyframes} $10 \%$ of the time steps (i.e., $n = N / 10$), uniformly
distributed in the time interval $[0, 1]$. The remaining $90 \%$ of the time
steps were considered as ground-truth data (not seen by the model during
training).
Specifically, we considered the following datasets.


\myparagraph{Gaussian mixture:} This synthetic 2D dataset counts $N = 180$ time
steps representing a mixture of six Gaussians, where the Gaussian centers
(captured by persistent maxima) evolve through time. Specifically, to
evaluate the robustness of
our method
to value changes as well as
feature displacements, this dataset is decomposed into
segments
where only the
weights of a few Gaussians evolve (i.e., making hills appear or disappear in
the corresponding terrain),
segments
where only the center positions for
a few Gaussians evolve (i.e., displacing hills in the corresponding terrain)
and
segments
where both phenomena occur.

\myparagraph{Mixing vortices:} This 2D dataset counts
$N = 150$
time steps
and represents the vorticity (measured as the orthogonal curl component) of a
2D flow generating a von Karman street (see \cite{ttkData} for downloads).
This dataset has the
particularity to model a 2D domain with boundaries. Then, in a first segment,
the vortices of the street (captured by persistent extrema) follow a typical,
common translation motion. However, in a second segment, the vortices
hit the boundary and consequently start to mix together in a complicated,
turbulent,
whirling pattern.

\myparagraph{Isabel:} This 3D dataset counts
$N = 48$
time steps representing
wind velocity for the Isabel hurricane \cite{scivis2004}.
To capture wind regions interacting with the boundary of the domain, we will
consider as input
field the opposite of the wind velocity. Then,
regions associated with high winds
will be
captured by persistent minima of this opposite velocity. In particular,
the eye of the hurricane
travels through the
domain into several stages (formation, drift,
and landfall).

\myparagraph{Asteroid impact:} This 3D dataset counts \todo{$N = 50$} timesteps
and represents the impact of an asteroid with the sea at the surface of the
Earth \cite{scivis2018}. The considered scalar field is matter density, which
distinguishes well the asteroid from the water and the ambient air in this
simulation. In this dataset, the trajectory of the asteroid as well as the
topological features resulting from the impact with the sea are captured from
a topological point of view by persistent maxima.

\begin{table}
  \caption{Meta-parameters adjusted empirically to
account for the variability in geometrical/temporal
complexity
across our
datasets.}
  \label{arch_configs}
  \adjustbox{width=\linewidth,center}{
    \begin{tabular}{|l|r|r|r|r|r|r|r|r|r|r|}
    \hline
    Dataset & $C_0$ & $C_1$ & $C_2$ &
    $C_3$ & $C_4$ &
    $C_5$ & $C_6$ & Learning rate & $n_1$ & $n_2$\\
    \hline
    Gaussian mixture (2D) & 256 & 128 & 64 & 64 & 32 & 16 & 8 & $0.5
\times 10^{-3}$ & 6000 & 100\\
    Mixing vortices (2D) & 256 & 128 & 64 & 64 & 32 & 16 & 8 & $0.5
\times 10^{-3}$ & 6000 & 100\\
    Isabel (3D) & 128 & 64 & 32 & 32 & 16 & -- & -- & $10^{-3}$ & 3000 & 100\\
    Asteroid impact (3D) & 512 & 256 & 128 & 64 & 16 & -- & -- & $10^{-4}$ &
3000 & 100\\
    \hline
    \end{tabular}
  }
\end{table}

\begin{figure}
\centering
\includegraphics[width=0.2425\linewidth]{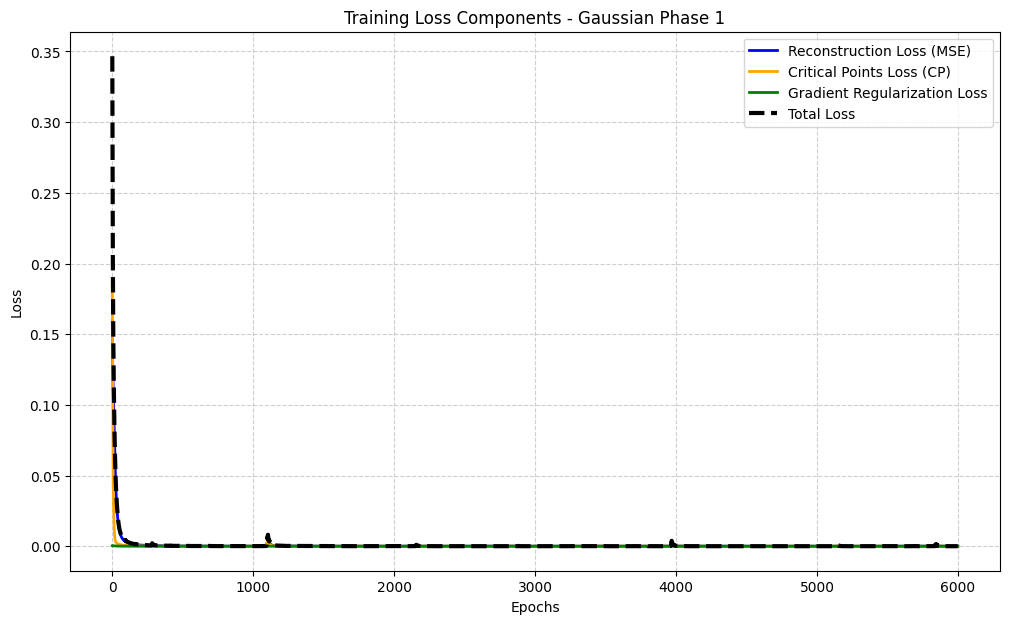}
\hfill
\includegraphics[width=0.2425\linewidth]{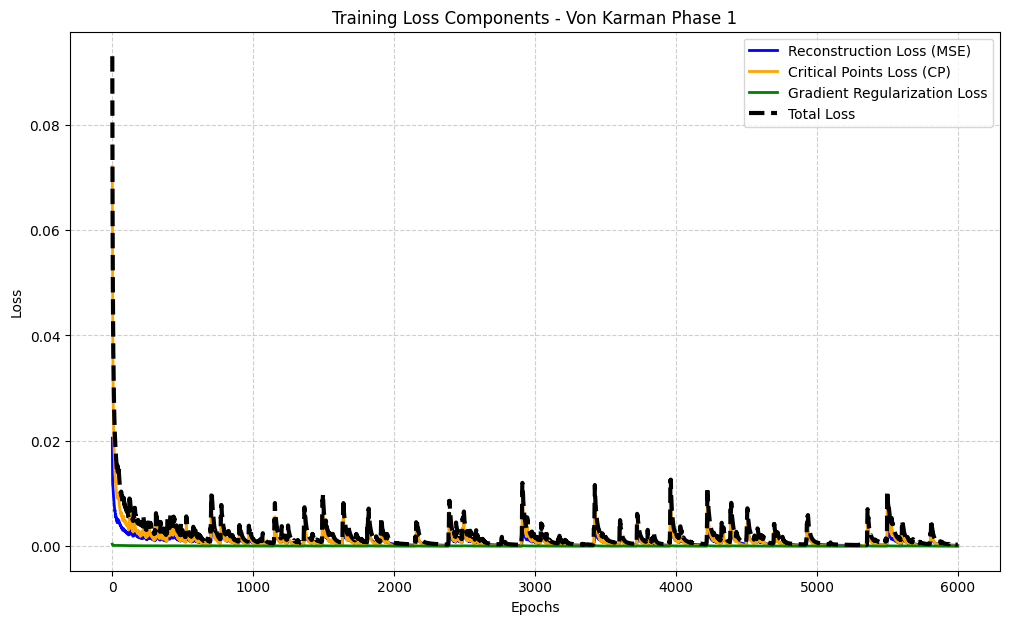}
\hfill
\includegraphics[width=0.2425\linewidth]{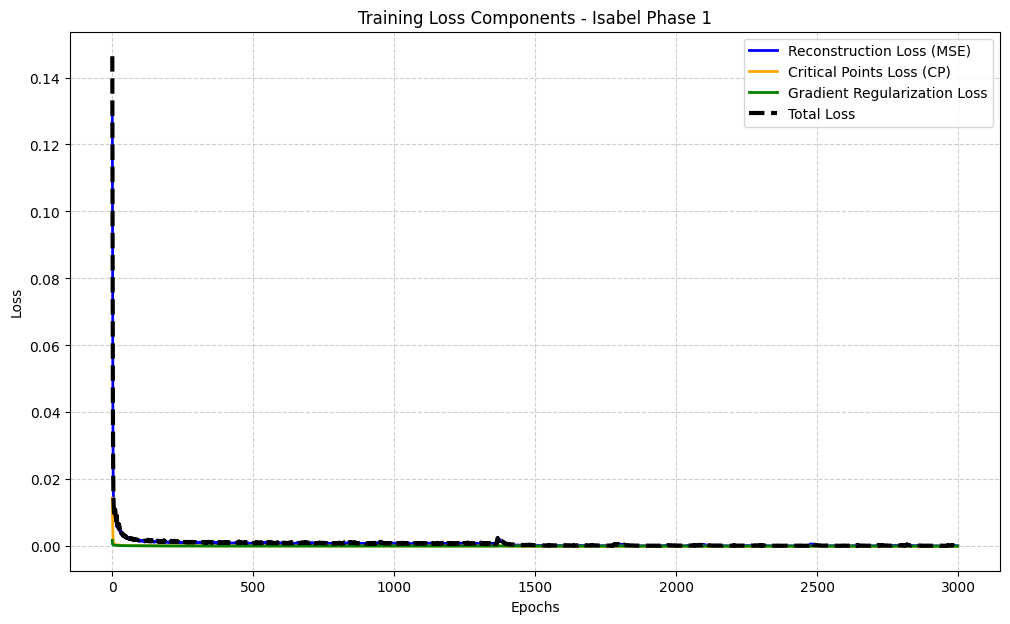}
\hfill
\includegraphics[width=0.2425\linewidth]{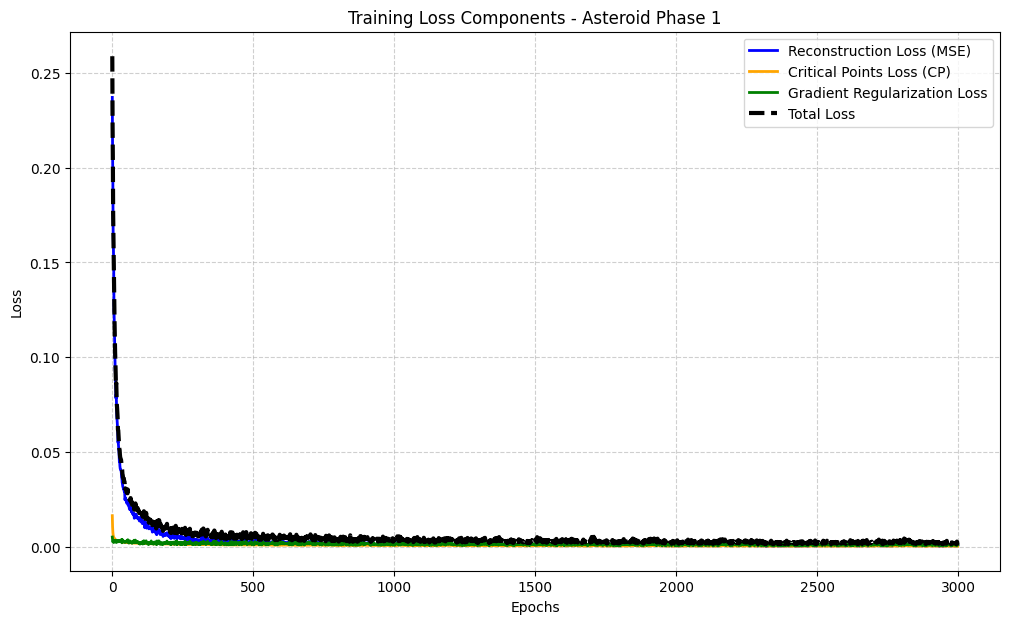}

~

\includegraphics[width=0.2425\linewidth]{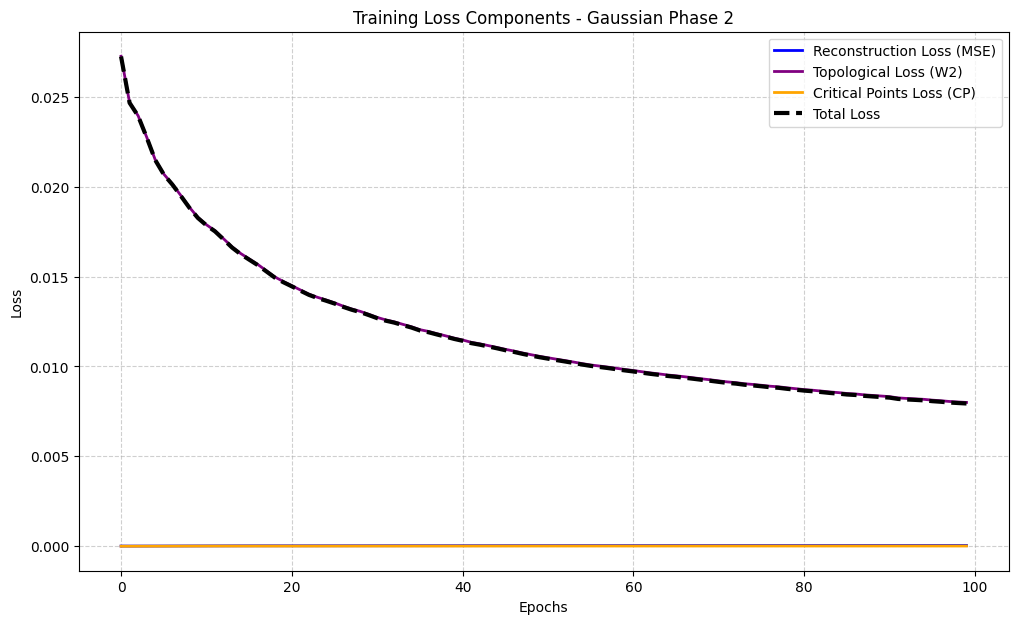}
\hfill
\includegraphics[width=0.2425\linewidth]{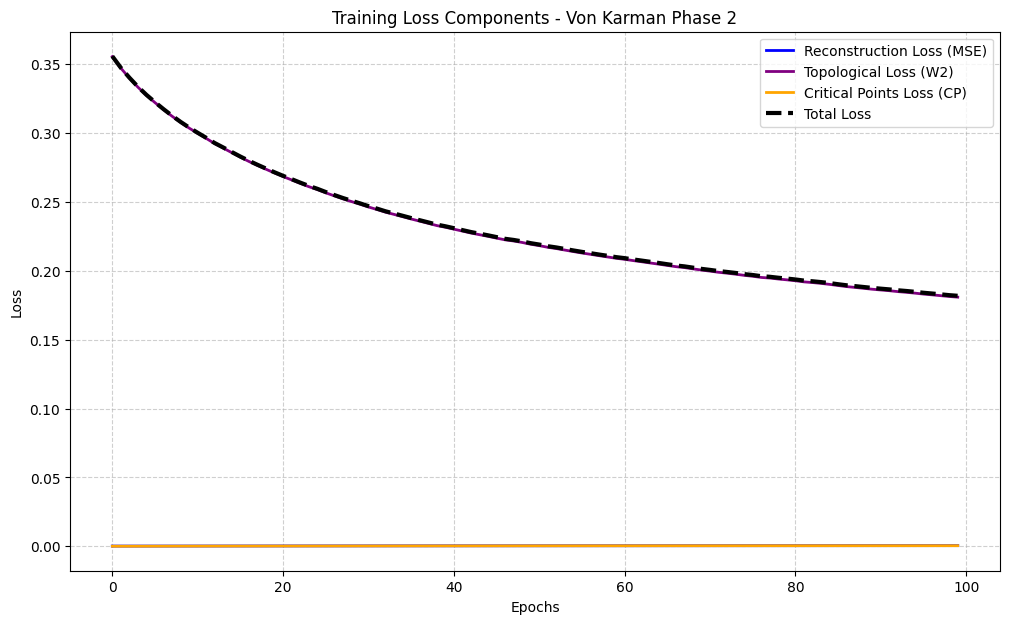}
\hfill
\includegraphics[width=0.2425\linewidth]{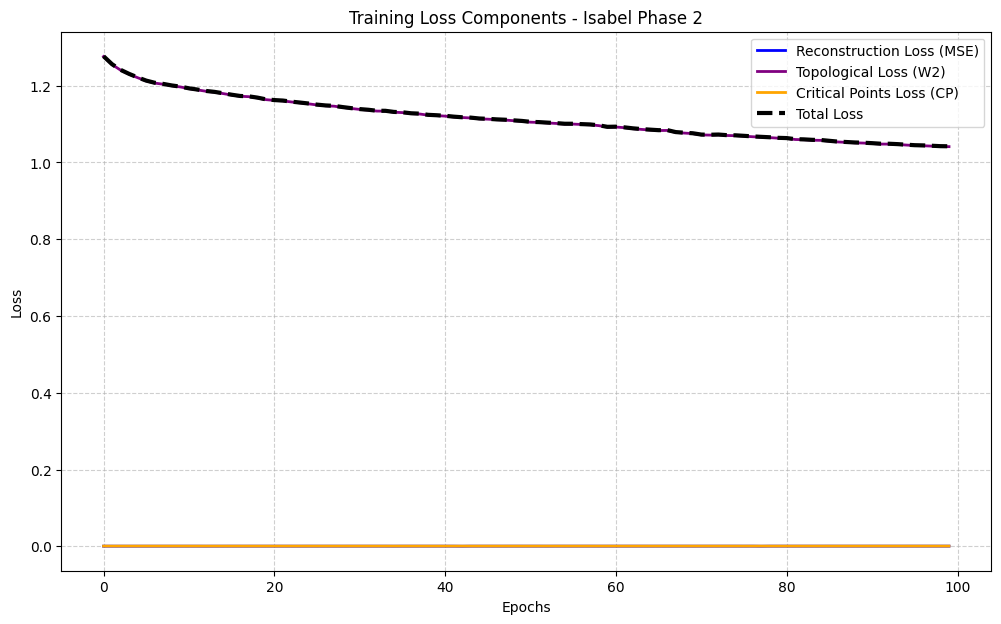}
\hfill
\includegraphics[width=0.2425\linewidth]{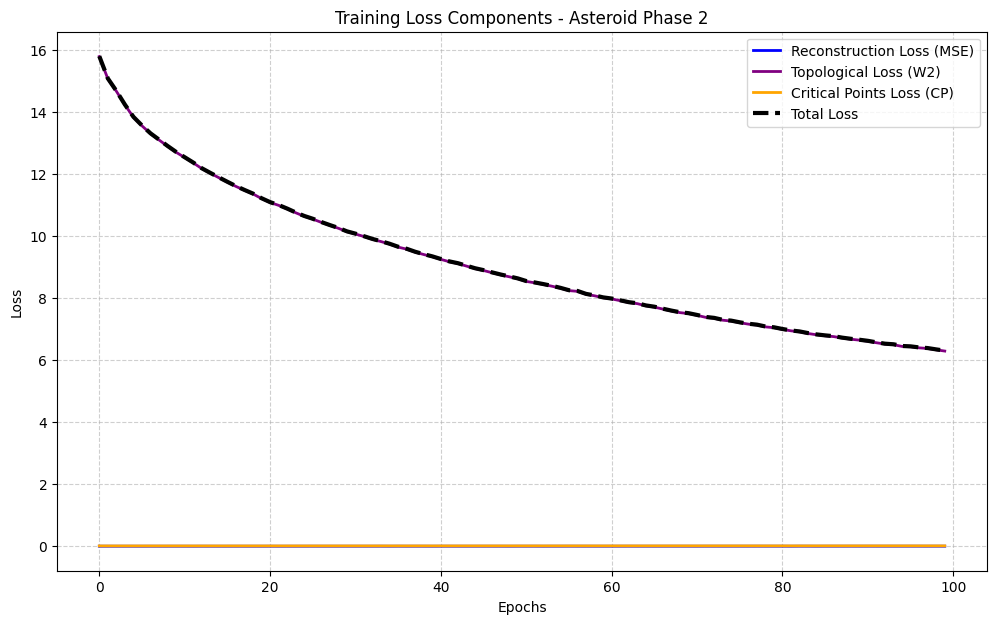}
\caption{Loss evolution during
\emph{scalar
field training} (phase $1$, top) and
\emph{topology correction} (phase
$2$, bottom) for our
datasets (from left to right: Gaussian mixture,
Mixing vortices, Isabel, Asteroid impact).}
\label{fig_convergenceCurves}
\end{figure}

Since these datasets exhibit distinct levels of details in terms of 
geometrical features and temporal variability, several 
meta-parameters of our approach were empirically adjusted on a per dataset
basis (\autoref{arch_configs}).
\autoref{fig_convergenceCurves} provides curves plotting the corresponding
loss, for each dataset.

\subsection{Reference approaches}
\label{sec_competition}
We compare our technique to two approaches,
selected based on
their query time, which is comparable to ours (\todo{typically,
below a second in practice}). First, we consider as
\topoinvis{a}
baseline the traditional
pointwise, linear interpolation with regard to time. This scheme  requires
the evaluation of a linear equation at each vertex of the input grid. Next,
among the
approaches introduced in the visualization
literature, we considered
STSR-INR \cite{tang2024stsr}, which is a recent representative of neural
methods
for spatio-temporal super resolution.
In particular, we used the implementation provided by the authors, set up to 
its default recommended parameters, except for:
\vspace{-1ex}
\begin{itemize}[itemsep=-.75mm]
  \item the epoch number (increased to obtain similar training times);
  \item the model size (increased to obtain sizes also similar to ours).
\end{itemize}
\vspace{-1ex}
The purpose of these modifications was to provide the same computational
resources to both methods (STSR-INR \cite{tang2024stsr} and ours).
Finally, note that in both cases (linear interpolation and STSR-INR),
these reference approaches have only access to the \emph{keyframe} data.

\subsection{Quantitative criteria}
We evaluate the quantitative quality of the resulting interpolations, for the
non-keyframes only, in the light of two fitting terms. First, we evaluate a
\emph{data fitting} term with the traditional
\emph{Peak Signal-to-Noise Ratio} (PSNR). Second, we evaluate a
\emph{topological fitting} term, which
measures
the topological accuracy of an
interpolation with regard to the input target diagram, based on the
\topoinvis{normalized}
Wasserstein
distance (\autoref{sec_wasserstein}).
For information, we also report indicators related to the computational
resources used in our experiments, namely model size (in MB) and training time
(in seconds).


\begin{figure}
\centering
\includegraphics[width=\linewidth]{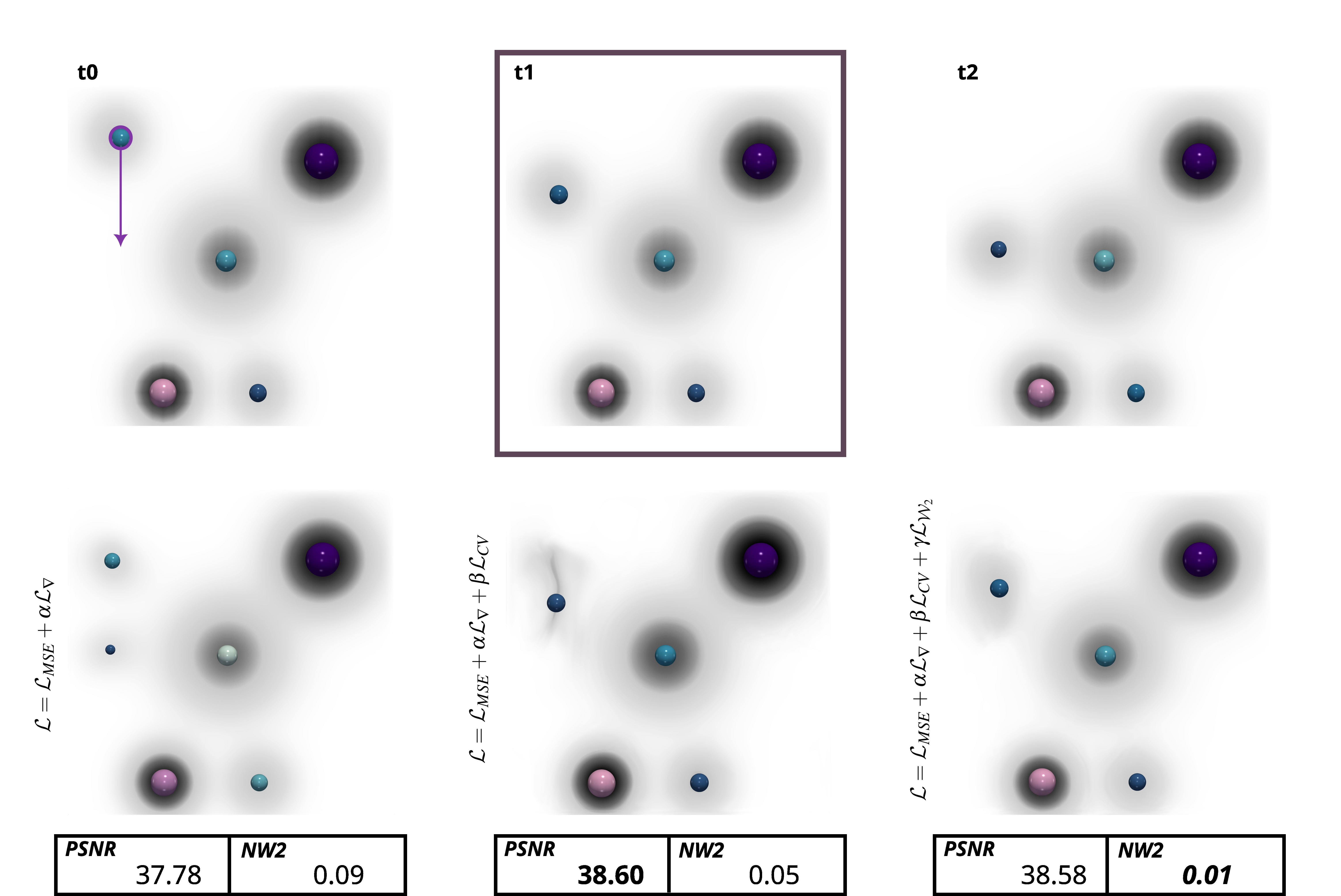}
\caption{%
Influence of our individual loss terms (\autoref{sec_losses}) on the
interpolation.
\emph{Top:} Three time steps of the \emph{Gaussian mixture} dataset (spheres:
local maxima, colored and scaled by persistence). From $t_0$ to $t_2$, a
Gaussian center is moving down, vertically (purple arrow, $t_0$).
\emph{Bottom:} Interpolations obtained for the time step $t_1$ with different
loss blending coefficients (\autoref{sec_details}), from left to right:
\todo{\emph{(i)} $\alpha = 0.1$, $\beta = \gamma = 0$ (no topological loss),
\emph{(ii)} $\alpha = 0.1$, $\beta = 1$, $\gamma = 0$ (critical value
enforcement),
\emph{(iii)} $\alpha = 0.1$, $\beta = 1$, $\gamma = 1$} (topology correction).
Without our topological losses (\emph{bottom, left}), the interpolation exhibits
the
\emph{superposition artifact} typical of linear interpolation, with the moving
Gaussian being replaced
by two Gaussians  of decreased
height (at the start and end points of the displacement). Critical value
enforcement
(\emph{bottom, center}) addresses this issue, resulting in one maximum in the
correct location, while topology correction (\emph{bottom, right}) improves the
final geometry and topology.
}
\label{fig_losses}
\end{figure}

\subsection{Loss influences}
\label{sec_influence}
We start the practical analysis of our method by investigating the effects of
the individual terms of our loss (presented in \autoref{sec_losses}).
\autoref{fig_losses} presents interpolation results on our synthetic dataset
(\emph{Gaussian mixture}) for various loss blending coefficients.
Specifically, this figure focuses on a temporal sequence where a Gaussian
center is moving down vertically (\revision{purple} arrow). When using only the
\emph{MSE} \todo{and gradient losses} (bottom, left), our method results in
interpolations with a \emph{superposition artifact} that is typical of linear
interpolation: the moving Gaussian has been replaced by two Gaussians of
decreased height (see the two maxima), at the start and end points of the
displacement. The introduction of our loss based on the critical values reported
by the input diagrams greatly contributes to addressing this issue (bottom,
center). Specifically, it results in a persistent maximum, at the right value
and at the right location.
\revision{However, the corresponding feature exhibits a ridge-like geometry 
(dark curve) which does not resemble the original feature (a Gaussian, 
top-center).}
The introduction of topology correction (based
on the Wasserstein distance to the input diagram\revision{, bottom right}) 
further improves topological
\revision{preservation (in terms of \topoinvis{normalized} Wasserstein
distance)},
as well as, importantly, the geometry of the prediction\revision{: the  
feature associated to the moving maximum exhibits a shape that is closer 
visually to the original Gaussian}.
\topoinvis{We assume that this geometry preservation is favored by the interplay
between the topology correction (which enforces the number and persistence of
features, irrespective of their location, \autoref{fig_persistenceDiagrams}) and
the critical value loss (which, by enforcing critical values at specific
locations, favors the appearance of topological features there).}

\begin{table}
\caption{Quantitative scores (averaged over all non-keyframe timesteps) over
our test datasets for the linear interpolation, STSR-INR \cite{tang2024stsr}
and our method. For each score, the best value is reported in
bold. Model sizes and training times are also reported.}
\label{table_quantitativeScores}
\adjustbox{width=\linewidth,center}{
  \begin{tabular}{|l|l|r|r|r|r|}
  \hline
  Criterion & Method & Gaussian Mixture & Mixing vortices &
Isabel & Asteroid Impact\\
  \hline
  ~ & \josh{Linear} Interpolation & 37.44 & 26.20 & 29.56 & 20.81\\
  PSNR ($\uparrow$)  & STSR-INR \cite{tang2024stsr} & 36.18 & 26.17 & 29.05 &
20.06\\
~
  & Our method & \textbf{38.58} & \textbf{29.28} &
\textbf{32.41}
& \textbf{22.51}\\
  \hline
  ~ & \josh{Linear} Interpolation & \topoinvis{0.12} & \topoinvis{0.40} & \topoinvis{0.62} &
\topoinvis{0.56}\\
  \topoinvis{$\normalizedWasserstein{2}$} ($\downarrow$) & STSR-INR
\cite{tang2024stsr} & \topoinvis{0.10} & \topoinvis{0.42}
& \topoinvis{0.60} & \topoinvis{0.67}\\
  ~
  & Our method & \textbf{\topoinvis{0.01}} & \textbf{\topoinvis{0.17}}
&
\textbf{\topoinvis{0.51}} & \textbf{\topoinvis{0.38}}\\
  \hline
  \hline
  \multirow{2}{*}{Size (MB)} & STSR-INR \cite{tang2024stsr} & 11 &
11 & 33 &
157\\
  ~ & Our method & 11 & 11 & 33 &
157\\
  \hline
  \multirow{2}{*}{Training \mohamed{(h.)}} & STSR-INR \cite{tang2024stsr} &
  \mohamed{3.33} & \mohamed{3.31} & \mohamed{6.76}
& \mohamed{9.69}\\
  ~
  & Our method & \mohamed{3.20} & \mohamed{2.76} &
  \mohamed{5.70} & \mohamed{8.96}\\
  \hline
  \end{tabular}
}
\end{table}

\begin{figure}
\centering
\includegraphics[width=\linewidth]{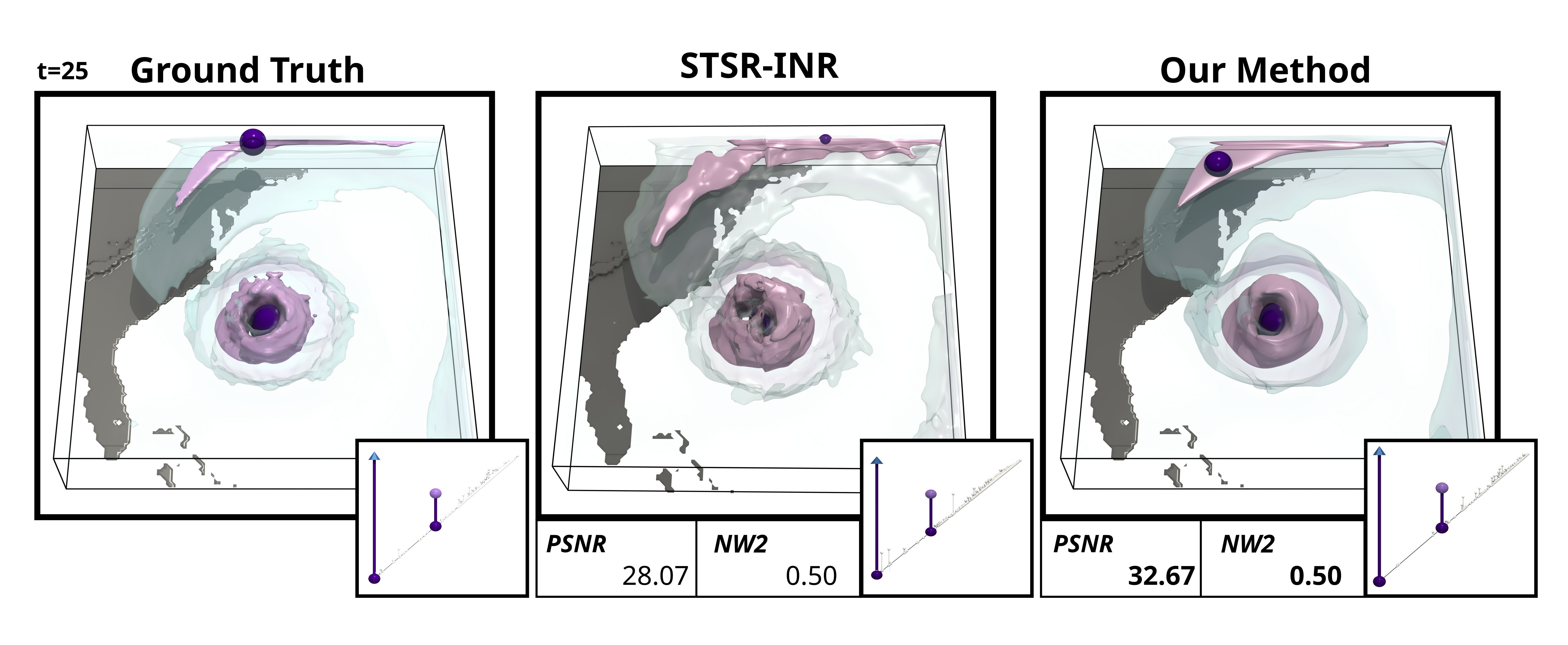}
\caption{
Comparing STSR-INR with a topological optimization post-process
\cite{kissi_vis24} (center) to the ground truth (left) and our method
(right).
To obtain a topological accuracy comparable to our method, the predictions
by STSR-INR need to undergo a
non-negligible optimization post-process
(56 seconds) while our method 
\revision{is instantaneous.}
%
}
\label{fig_postProcess}
\end{figure}

\subsection{Comparisons}
This section provides a quantitative and qualitative comparison between our
method and
reference approaches
(\autoref{sec_competition}).

\autoref{table_quantitativeScores} provides an overview of the quantitative
scores (averaged over all non-keyframe time steps) over our test datasets for
the considered methods. This table shows that our method, by design, provides
the best topological accuracy (\topoinvis{i.e., normalized} Wasserstein
distance to the
input diagrams).
As discussed in \autoref{sec_influence}, our topological losses
(\autoref{sec_losses}), on top of contributing to
\revision{the preservation of topological features,}
also contribute 
to improving the
geometry of the prediction. This is illustrated by the PSNR scores
of
our method, which are superior to those of the reference approaches.

\begin{figure*}
\centering
\includegraphics[width=\linewidth]{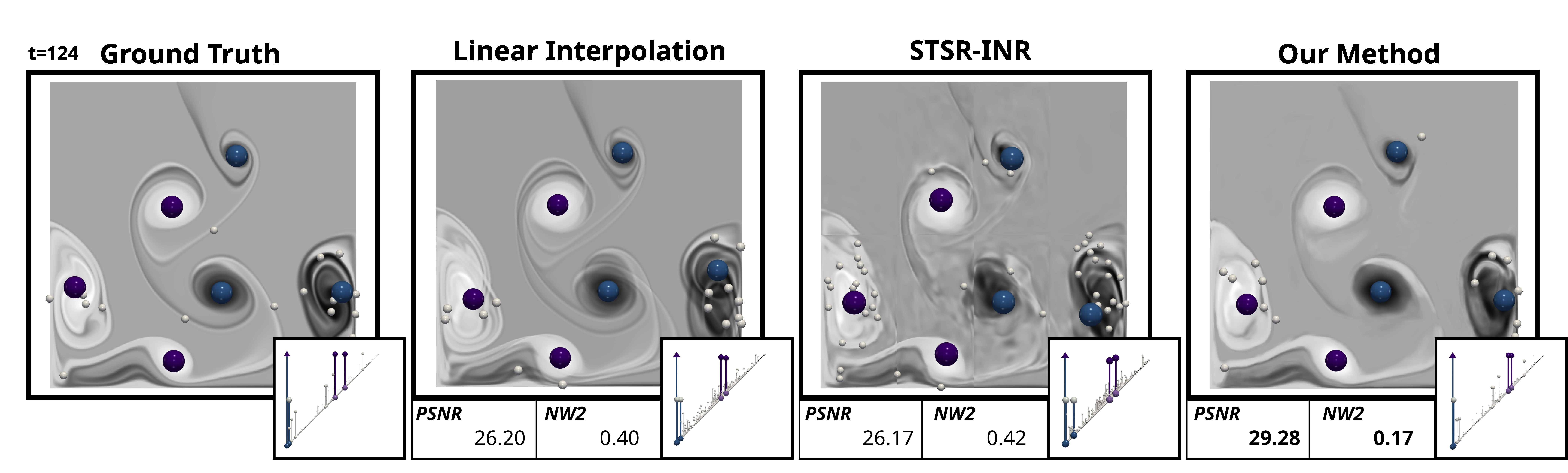}
\caption{%
Comparison of temporal interpolations on the \emph{Mixing vortices} dataset
(from left to right: ground-truth, linear interpolation, STSR-INR
\cite{tang2024stsr} and our method). Persistent extrema are reported as colored
spheres (blue: minima, purple: maxima), extrema of intermedia persistence are
reported as white spheres. The persistence diagrams are reported in the bottom
insets. The linear interpolation exhibits a typical \emph{superposition
artifact}, where the two keyframes used for the interpolation are superposed on
top of each other. This artifact is addressed by STSR-INR, with an improved
PSNR. Our method further improves PSNR, while improving topological
accuracy (in particular with fewer noisy bars in the diagram).
}
\label{fig_comparison_vortices}
\end{figure*}

\begin{figure*}
\centering
\includegraphics[width=\linewidth]{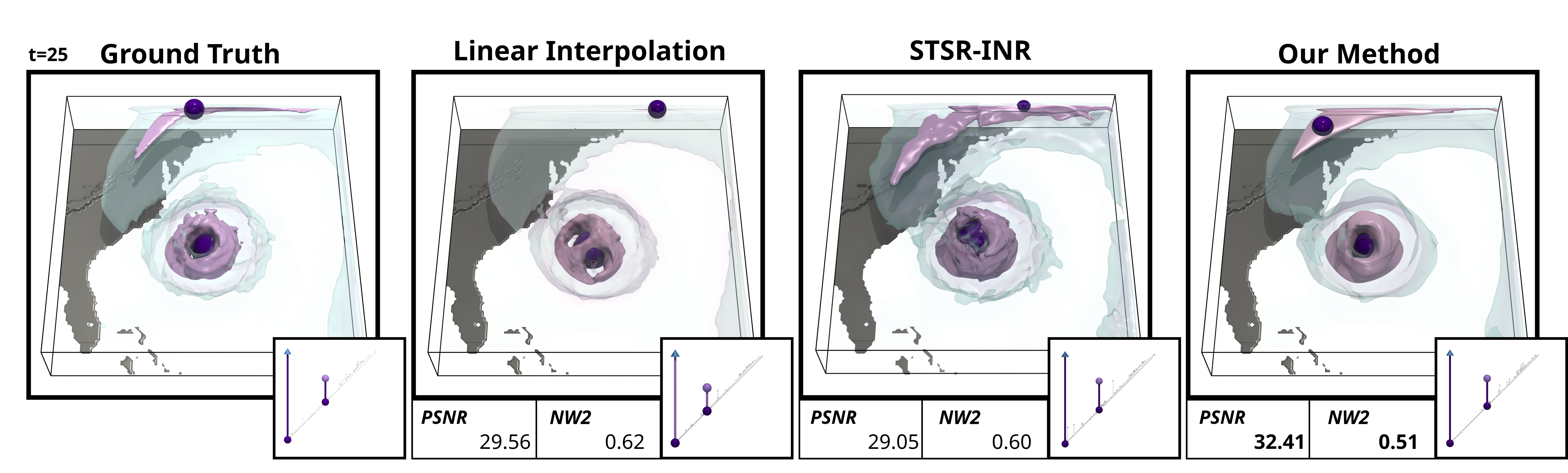}
\caption{%
Comparison of temporal interpolations on the \emph{Isabel} dataset
(from left to right: ground-truth, linear interpolation, STSR-INR
\cite{tang2024stsr} and our method). Persistent minima of the opposite wind
velocity are reported as purple
spheres. A few isosurfaces are shown
to represent the geometry of the data.
The persistence diagrams are reported in the bottom
insets. The linear interpolation
exhibits  its typical \emph{superposition
artifact} (similarly to \autoref{fig_comparison_vortices}), where the hurricane
eyes from two keyframes are superposed. Moreover, it
fails at capturing certain high wind regions
(light purple surface, top). Here, STSR-INR produces a reconstruction
with a degraded PSNR.
Our method provides the best PSNR and the best topological
accuracy.
}
\label{fig_comparison_isabel}
\end{figure*}

\begin{figure*}
\centering
\includegraphics[width=\linewidth]{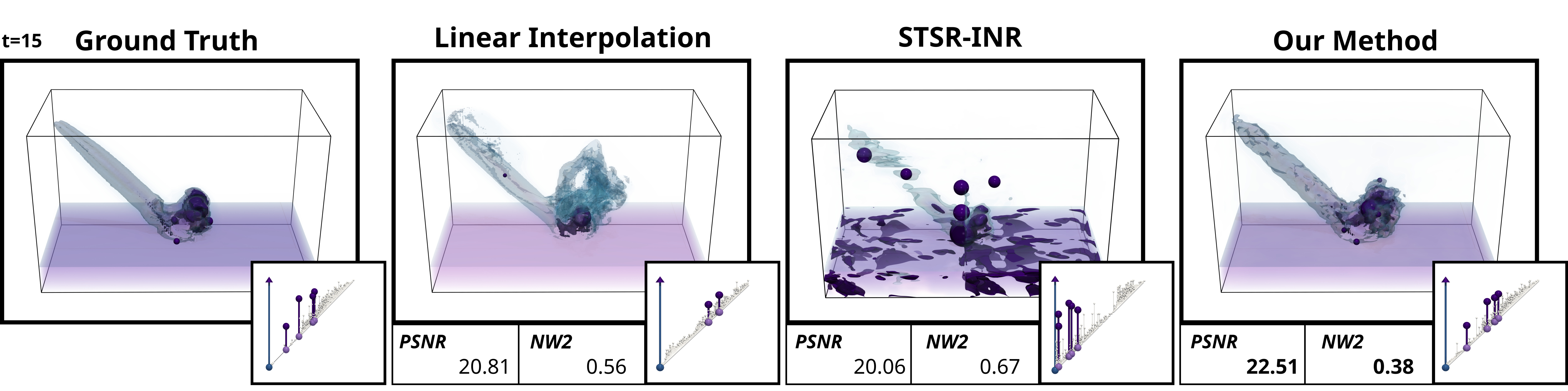}
\caption{%
Comparison of temporal interpolations on the \emph{Asteroid impact} dataset
(from left to right: ground-truth, linear interpolation, STSR-INR
\cite{tang2024stsr} and our method). Persistent maxima are reported as purple
spheres. The geometry of the data is represented via volume rendering and
isosurfacing.
The persistence diagrams are reported in the bottom
insets.
The linear interpolation exhibits its typical \emph{superposition
artifact},  where the two keyframes used for the interpolation are superposed on
top of each other
(similarly to \autoref{fig_comparison_vortices}).
Here,
STSR-INR produces a
result
with a degraded PSNR.
Our method provides the
indicators
scores
(PSNR and topological
accuracy),
and it provides
the result which conforms best
visually
to the ground truth.
}
\label{fig_comparison_asteroid}
\end{figure*}

Qualitative comparisons are provided in Figs. \ref{fig_comparison_vortices},
\ref{fig_comparison_isabel}, \ref{fig_comparison_asteroid}.
Overall, these figures show that the linear interpolation suffers from
a typical \emph{superposition artifact}, where the two keyframes used for the
interpolation are superposed on
top of each other.
This is particularly severe when features of interest are moving
within the domain across time (which occurs in all our test datasets).
This is partly addressed, sometimes successfully
(\autoref{fig_comparison_vortices}), sometimes less successfully
(\autoref{fig_comparison_asteroid}), by the STSR-INR \cite{tang2024stsr}
approach, which improves PSNR in some case (\autoref{fig_comparison_vortices}).
In all cases, our approach better
preserves, by design, the topological features, resulting in a superior
topological accuracy. 
\topoinvis{In}
several instances (Figs. \ref{fig_comparison_isabel},
\ref{fig_comparison_asteroid}),
our method
provides the result which
conforms best visually
\topoinvis{(geometrically)}
to the
ground truth \revision{(confirming the observations of 
\autoref{sec_influence})}.

In principle, the results of the reference approaches
\revision{(e.g., STSR-INR
\cite{tang2024stsr}),}
could be post-processed to improve
their
topological accuracy. This can be done, by example, by using the approach by
Kissi et al. \cite{kissi_vis24}.
However, as shown in \autoref{fig_postProcess}, such a post-processing step is
non-negligible in terms of runtime, \revision{which} significantly degrades  
interpolation
query response
times.
In contrast, at query time, our approach provides  
\revision{a result with comparable topological accuracy instantaneously.}

%


\subsection{Limitations}
\label{sec_limitations}

An obvious limitation of our work, which is intrinsic to neural approaches in
general (such as STSR-INR \cite{tang2024stsr}), is the computational effort
required for training such models,
in the range of hours of computation
(\autoref{table_quantitativeScores}).
However, for applications such as in-situ computing
\cite{insitu, AyachitBGOMFM15}, for which data storage can be a more important
concern than computational effort, we believe our
data reduction
strategy to be still relevant, as assessed by our
model sizes
(\autoref{table_quantitativeScores}).

Another limitation that we observed was the need for
larger
models for the datasets exhibiting the most geometrical and temporal complexity
(Tabs. \ref{arch_configs}, \ref{table_quantitativeScores}). While it is
understandable that more parameters are required in the model to capture this
variability, this model size increase negatively impacts training computation
times as well as
model storage.
\revision{Similarly, larger output sizes may require
more
iterations, themselves being more computationally expensive (each iteration
involves \topoinvis{a computation of persistence).}}
Finally, our approach is currently restricted to regular grids,
which
is the only
representation supported by our CNN-based decoding.
Alternative generative architectures would need to be considered for more
generic inputs, e.g., scalar fields defined on
tetrahedral meshes.

%
%
%

\section{Conclusion}
\label{sec_conclusion}

This paper presented a neural approach for the topology aware interpolation of
scalar fields. Our work was motivated by a data model often encountered in
in-situ computing \cite{insitu, AyachitBGOMFM15}, where snapshots
of the considered time-varying data are only saved at a low frequency (every $n$
\emph{keyframes}) and where reduced representations, such as topological
descriptors, are stored at a higher frequency \cite{bremer_tvcg11,
BrownNPGMZFVGBT21, vestec} (every $N \gg n$ steps). Specifically, given an
input sequence of persistence diagrams and a sparse temporal sampling of the
corresponding data,
our approach \emph{``inverts''} the non-keyframe diagrams to produce plausible
estimations of the
missing data.
Extensive experiments showed the superiority of our method over reference
approaches for preserving the topological features of interest along the
interpolation. Interestingly, our experiments also revealed that our
topology-aware losses could also 
contribute to improving the
\emph{geometry} of the interpolated data.

\topoinvis{In the future, we will extend our experimental
evaluations, by considering additional datasets and quality metrics
\cite{tang2024stsr}.}
We
believe our work opens several research avenues, in particular, thanks to
its instantaneous query time, for
the interactive exploration of ensembles of topological descriptors, as studied
in topology-tailored statistical frameworks \cite{Turner2014, lacombe2018,
YanWMGW20, pont_vis21,
pont_tvcg23, pont_tvcg24}. However, a multidimensional extension of
approach (to account for more ensemble parameters than simply time) would need
to be investigated.



\acknowledgments{%
\topoinvis{
\footnotesize{This work is partially supported by the European Commission
grant ERC-2019-COG \emph{``TORI''} (ref. 863464,
\url{https://erc-tori.github.io/}),
by the U.S. Department of Energy, Office of Science, under Award Number(s)
DE-SC-0023319, and by a joint graduate research fellowship (ref. 320650) funded
by the CNRS and the University of Arizona.}}}

\bibliographystyle{abbrv-doi-hyperref}

\bibliography{template,josh}


\end{document}